\documentclass[twocolumn,10pt]{asme2ej}
\usepackage{amsmath,bm} % assumes amsmath package installed
\usepackage{mathtools}
\usepackage{amssymb}  % assumes amsmath package installed
\usepackage{tabularx}
\usepackage{xfrac}
\usepackage{booktabs}
\usepackage{graphicx}
\graphicspath{{../Ch4/}}

\title{The Right Invariant Nonlinear Complementary Filter for Low Cost Attitude and Heading Estimation of Platforms}

\author{Oscar De Silva
    \affiliation{
   Assistant Professor\\
	Faculty of Engineering-\\and Applied  Science\\
Memorial University of Newfoundland\\
St. John's, NL, Canada. \\
    Email: oscar.desilva@mun.ca
    }	
}

\author{George K. I. Mann
    \affiliation{
    Professor\\
	Faculty of Engineering and-\\Applied  Science\\
Memorial University of Newfoundland\\
St. John's, NL, Canada. \\
    Email: gmann@mun.ca
    }	
}

\author{Raymond G. Gosine
    \affiliation{
    Professor\\
	Faculty of Engineering and Applied  Science\\
Memorial University of Newfoundland\\
St. John's, NL, Canada. \\
    Email: rgosine@mun.ca
    }	
}

\begin{document}

\maketitle    

%%%%%%%%%%%%%%%%%%%%%%%%%%%%%%%%%%%%%%%%%%%%%%%%%%%%%%%%%%%%%%%%%%%%%%
\begin{abstract}
{\it This paper presents a novel filter with low computational demand to address the problem of orientation estimation of a robotic platform. This is conventionally addressed by extended Kalman filtering of measurements from a sensor suit which mainly includes accelerometers, gyroscopes, and a digital compass. Low cost robotic platforms demand simpler and computationally more efficient methods to address this filtering problem.  Hence nonlinear observers with constant gains have emerged to assume this role. The nonlinear complementary filter is a popular choice in this domain which does not require covariance matrix propagation and associated computational overhead in its filtering algorithm. However, the gain tuning procedure of the complementary filter is not optimal, where it is often hand picked by trial and error. This process is counter intuitive to system noise based tuning capability offered by a stochastic filter like the Kalman filter. This paper proposes the right invariant formulation of the complementary filter, which preserves Kalman like system noise based gain tuning capability for the filter. The resulting filter exhibits efficient operation in elementary embedded hardware, intuitive system noise based gain tuning capability and accurate attitude estimation. The performance of the filter is validated using numerical simulations and by experimentally implementing the filter on an ARDrone 2.0 micro aerial vehicle platform.
}
\end{abstract}

%%%%%%%%%%%%%%%%%%%%%%%%%%%%%%%%%%%%%%%%%%%%%%%%%%%%%%%%%%%%%%%%%%%%%%
%\begin{nomenclature}
%\entry{A}{You may include nomenclature here.}
%\entry{$\alpha$}{There are two arguments for each entry of the nomemclature environment, the symbol and the definition.}
%\end{nomenclature}
%
%The primary text heading is  boldface and flushed left with the left margin.  The spacing between the  text and the heading is two line spaces.

%%%%%%%%%%%%%%%%%%%%%%%%%%%%%%%%%%%%%%%%%%%%%%%%%%%%%%%%%%%%%%%%%%%%%%
\section{Introduction}

% The problem of attitude estimation and the general solution
The objective of an Attitude and Heading Reference System (\textbf{AHRS}) is to estimate the orientation of a platform using measurements from multiple sensors. Typically these sensors may include accelerometers, magnetometers, gyroscopes, and velocity sensors.  Orientation estimation of robots using these measurements is traditionally performed using the Extended Kalman Filter (\textbf{EKF}) \cite{Lefferts1982}. The EKF is a natural choice for this application due to its popularity in many similar nonlinear estimation problems. Notable applications related to robotics are localization \cite{Roumeliotis2002, Silva2014}, simultaneous localization and mapping \cite{Thrun2005a}, target tracking \cite{BlackmanS.S.andPopoli1999}, and inertial navigation systems \cite{Farrell2008a}. 

 % The problems associated with linear error state assumptions
Despite its popularity in many applications, the classical EKF has several drawbacks. The EKF assumes a linear state error and a linear output error in its formulation. This is an inaccurate error representation for many practical applications including the orientation estimation problem.  The classical EKF formulation tends to overlook the underlying geometry related to the rotation group \cite{Mahony2008}. Hence the EKF is required to maintain redundant states to represent rotation errors and those states propagate in a geometrically inconsistent manner during operation. This issue has been effectively addressed by nonlinear filters where filtering equations and error definitions are consistent with the rotation group. The nonlinear filter designs presented in literature can be broadly classified into two major groups, either as deterministic filters or as stochastic filters.
 
 %results in solutions with similar or reduced computation demand in comparison to an EKF, but with increased difficulty in systematic design for a general system. 
 
 % Nonlinear filters for low cost applications and their drawbacks
Deterministic filter formulations do not consider the random noise variables as well as the covariance figures related to those noise variables. As a result, these filters  have a minimal set of filtering equations and parameters (gains). The Nonlinear Complementary Filter (\textbf{NCF}) is one of the popular deterministic forms available for attitude and heading estimation of robotic platforms and radio controlled platforms \cite{Hamel2006}. It has global stability implications and low computational requirement\cite{Mahony2008} .  One of the major drawbacks of NCF is that there is no systematic procedure available to tune its gains, where the gains are usually selected by trial and error methods. %In contrast,  a stochastic filter like an EKF possesses quite intuitive and direct means of tuning gains while exploiting the measurement noise figures of the sensors. %Hence not having a systematic gain tuning procedure can be considered as a major drawback of the NCF \cite{DeSilva2013}.

%Stochastic nonlinear filters and drawbacks
Nonlinear stochastic filters consider random noise variables and their covariance figures in its formulation. The gains of such stochastic filters are computed adaptively based on system noise statistics. This capability provides better convergence properties and intuitive tuning properties which are not available in deterministic forms (e.g. NCF). One of the major drawbacks of nonlinear stochastic filters is the higher computational requirement needed for propagation of noise covariance values and calculation of adaptive gains. A popular nonlinear stochastic filter used for AHRS is the Multiplicative Extended Kalman Filter (\textbf{MEKF})  \cite{MARKLEY}. The MEKF introduces multiplicative error state definitions when computing rotational errors. These are more accurate than the linear error state definitions used by the EKF and this modification has become quite standardized in filter design tasks when rotational dynamics are involved \cite{Farrell2008a}.  A more recent breakthrough in AHRS design domain is the symmetry preserving observer design method \cite{Bonnabel2008}. This method is a generalized approach which allows to systematically  design filters for problems possessing symmetries such as robot localization \cite{Hervier2012, Bonnabel2012, DeSilva2014a}, inertial navigation systems \cite{Barczyk2013}, and AHRS \cite{Bonnabel2009, Martin2010}. The error system produced by following the symmetry preserving filter design approach can be stabilized using an error state Kalman filter as shown in \cite{Martin2009}. The resulting  nonlinear filter is termed as the Invariant Extended Kalman Filter (\textbf{IEKF}). A remarkable characteristic of the IEKF is its capability to converge its adaptive gains, to a constant set of values when the filter is operating along a class of trajectories termed ``permanent trajectories'' \cite{Martin2009}. In practice it is found that typical smooth trajectories of rotating platforms often operate in a domain of permanent trajectories where constant gains of the filter are observed during steady state operations \cite{Martin2009}.

%Implications from recent results
Therefore, one may raise the question as to whether the NCF is actually a steady state version of the IEKF (or whether there is any significant connection) because the IEKF tends to behave like a constant gain filter during steady state operation. This is an important identification since it would imply the possibility of using steady state gains of the IEKF to tune the NCF. Since the gains of the IEKF are parameterized by the systems' noise parameters, the resulting gain computation method would be quite intuitive and systematic for practical applications. This paper exploits the above idea to realize a novel formulation of the NCF which can be tuned using the steady state converged gain values of an equivalent stochastic formulation of the filter. 

The paper first uses a symmetry preserving observer design approach to identify a stochastic equivalent corresponding to the NCF.  Then a tuning method for the right invariant formulation of the NCF is proposed which uses the steady state gains of its stochastic equivalent. These steady state gains can be calculated using Discrete Algebraic Riccati Equation (\textbf{DARE}) solve functionality \cite{Arnold1984}, which is available in many mathematical packages. The work further discusses the stability implications of the design and validates the  proposed method using experimental evaluation on a Micro Aerial Vehicle (\textbf{MAV}).

%objective of this work and contribution
 This paper has three main contributions. First, this paper proposes a novel low cost filter for orientation estimation. This filter is termed as the Right Invariant Nonlinear Complementary Filter (\textbf{RINCF}). Second, this paper proposes a novel tuning method for the filter while exploiting the sensors' noise parameters. The tuning method produces gain values for the RINCF which are comparable to the gains of a steady state stochastic version of the filter. Third, this paper presents a detailed experimental evaluation of the filter when deployed on a typical hobby grade MAV device validating its capability to operate with minimal computational overhead.

% The most popular derivative of the EKF is the Multiplicative Extended Kalman Filter. This filter assumes a more accurate multiplicative error term for the state errors. The result is a filter more accurate error system models []. This flavour of the attitude estimator is used in many applications including attitude estimation and inertial navigation. However the filter is considered rather a trick used to counter nonlinearities encountered in rotation dynamics. This filter was only realized in a systematic framework recently in the work of []. 
 
 %This work present a method to realize  nonlinear filters for systems which exhibit symmetry. The resulting nonlinear filtering equations by following this approach can be satbilized by following an error state Kalman filtering approach. This filter is termed the Invariant Extended Kalman Filter and results in more geometrically sound filtering equations for the attitude estimation problem

%%%%%%%%%%%%%%%%%%%%%%%%%%%%%%%%%%%%%%%%%%

\section{Preliminaries}

\subsection{System and measurement model}

This work uses a quaternion parameterization $\mathbf{q}=(a~b~c~d)^{\mathsf{T}}$ to represent platform rotations from an earth aligned frame $\{E\}$ to a frame attached to the rotating body $\{B\}$. The multiplication operation between two quaternions $\mathbf{q}_1=(a_1~b_1~c_1~d_1)^{\mathsf{T}}$ and $\mathbf{q}_2=(a_2~b_2~c_2~d_2)^{\mathsf{T}}$ is defined using the operator $*$ as follows:
%\mathbf{q}_1=(a_1~b_1~c_1~d_1)^{\mathsf{T}}$ and $\mathbf{q}_2=(a_2~b_2~c_2~d_2)^{\mathsf{T}}$
\begin{equation*}
\mathbf{q}_1*\mathbf{q}_2=\begin{pmatrix}
a_1a_2 - b_1b_2 - c_1c_2 - d_1d_2\\
a_1b_2 + b_1a_2 + c_1d_2 - d_1c_2\\
a_1c_2 - b_1d_2 + c_1a_2 + d_1b_2\\
a_1d_2 + b_1c_2 - c_1b_2 + d_1a_2\\
\end{pmatrix}
\end{equation*}

All vectors $\mathbf{v}\in \mathbb{R}^3$ are augmented as $\mathbf{v}=(0 ~\mathbf{v}^{\mathsf{T}})^{\mathsf{T}}$ when performing a quaternion multiplication. The matrix $\mathbf{R}_q$ denotes the rotation matrix  corresponding to platform rotations from earth aligned frame $\{E\}$ to a frame attached to the rotating body $\{B\}$ which is parameterized by $\mathbf{q}$. I.e., $\mathbf{R}_q \in \mathbb{SO}(3)~~s.t.~\mathbf{q}*\mathbf{v}*\mathbf{q}^{-1}=\mathbf{R}_q\mathbf{v}~~,\forall \mathbf{v} \in \mathbb{R}^3$. 

%The subscript $q$ denotes that it is  parameterized by the quaternion $\mathbf{q}$ as given below.
%\begin{equation*}
%\mathbf{R}_q=\begin{pmatrix}
%a\\
%b\\
%c\\
%d
%\end{pmatrix}
%\end{equation*}
Following a quaternion parameterization, the low cost attitude and heading estimation problem faced by robotic platforms can be expressed using the system and measurement models given in equations (\ref{eqn4:qsys_model}) and (\ref{eqn4:qmeas_model}).

\begin{equation}\label{eqn4:qsys_model}
       \displaystyle  \begin{array}{c c}
        \dot{\mathbf{x}} = f(\mathbf{x},\mathbf{u},\mathbf{w})\\
        \begin{array}{c c c}
            \begin{pmatrix}
                \dot{\mathbf{q}}\\
                \dot{\bm{\omega}}_b\\
            \end{pmatrix}
                &=
            \begin{pmatrix}
               \frac{1}{2}\mathbf{q}*(\boldsymbol{\omega}_m-\boldsymbol{\omega}_b + \mathbf{w}_{\omega_m})\\
               \mathbf{w}_{\omega_b}
            \end{pmatrix}
        \end{array}
        \end{array}
    \end{equation}

The system model (\ref{eqn4:qsys_model})  has a state vector denoted by $\mathbf{x}:=(\mathbf{q}^{\mathsf{T}},~\boldsymbol{\omega}_b^{\mathsf{T}})^{\mathsf{T}}$, a process noise vector denoted by $\mathbf{w}:=(\mathbf{\mathbf{w}}_{\omega_m}^{\mathsf{T}},~\mathbf{\mathbf{w}}_{\omega_b}^{\mathsf{T}})^{\mathsf{T}}$, and an input vector denoted by $\mathbf{u}:=\boldsymbol{\omega}_m$. The vector $\boldsymbol{\omega}_m$ denotes a 3 axis gyroscope measurement which measures the absolute angular rate of the platform expressed in body fixed frame $\{B\}$. The gyroscope measurement has a bias term $\boldsymbol{\omega}_b$, and is corrupted by a measurement noise term $\mathbf{\mathbf{w}}_{\omega_m}$ along all three axis. The gyroscopic bias term follows a random walk process with noise $\mathbf{\mathbf{w}}_{\omega_b}$. 

 \begin{equation}\label{eqn4:qmeas_model}
        \begin{array}{c c}
        \mathbf{y} = g(\mathbf{x},\boldsymbol{\nu})\\
        \begin{array}{c c c}
            \begin{pmatrix}
                \mathbf{y}_a\\
                \mathbf{y}_b\\
            \end{pmatrix}
                &=
            \begin{pmatrix}
               -\mathbf{R}_{q}^\mathsf{T}\mathbf{g}_e+ \boldsymbol{\nu}_{a}\\
               \mathbf{R}_{q}^\mathsf{T}\mathbf{b}_e+ \boldsymbol{\nu}_{b}
            \end{pmatrix}
        \end{array}
        \end{array}
    \end{equation}

The measurement model (\ref{eqn4:qmeas_model}) has a measurement vector denoted by $\mathbf{y}:=(\mathbf{y}_a^{\mathsf{T}}, ~ \mathbf{y}_b^{\mathsf{T}})^{\mathsf{T}}$ and a measurement noise vector denoted by  $\pmb{\nu}:=(\pmb{\nu}_a^{\mathsf{T}}, ~ \pmb{\nu}_b^{\mathsf{T}})^{\mathsf{T}}$. Vector $\mathbf{y}_a$ denotes the measurements of a 3 axis accelerometer which measures gravitational and inertial accelerations of the platform expressed in body fixed frame $\{B\}$. Vector $\mathbf{g}_e$ denotes the gravitational reference vector expressed in frame $\{E\}$. For low cost attitude estimation it is assumed that platform inertial accelerations are absorbed in the accelerometer measurement noise term given by $\boldsymbol{\nu}_a$. As a result, accelerations due to gravity dominates the accelerometer measurement. Vector $\mathbf{y}_b$ denotes the measurements of a 3 axis magnetometer attached to the platform which measures the magnetic field intensity of the environment expressed in body fixed frame $\{B\}$. Vector $\mathbf{b}_e$ denotes the earth's magnetic reference expressed in frame $\{E\}$. The magnetometer model assumes that it measures the earth's magnetic field while all other disturbances are absorbed in the magnetometer measurement noise term given by $\boldsymbol{\nu}_a$.  During indoor navigation the magnetometer is given less confidence since it is heavily prone to errors produced by local magnetic disturbances \cite{Suksakulchai2000}.

\subsection{Symmetry preserving observer design approach}

The proposed filter in this paper exploits the filter design process of \cite{Bonnabel2008} to achieve a steady state deterministic implementation which is capable of incorporating system noise parameters for tuning purposes. The design procedure follows a differential geometric formulation. Important definitions and necessary notation, which are relevant to this work are summarized here.  Only the definitions that are necessary to follow the modified observer design process proposed in this work are presented in this section. All definitions and theorems related to a standard symmetry preserving observer design process are  discussed in detail in the original work of Bonnabel et al. \cite{Bonnabel2008}. 

\vspace{0.1cm}
\noindent \emph{\textbf{Definition 1}}: A set $G$ is a \emph{Group}, if for all $g\in G$, we can define  a \emph{Group operation} (multiplication) $\diamond$ with a corresponding identity element $e$, and an inverse element $g^{-1}$, which satisfies the conditions of closure and associativity.

As an example, unit quaternions $\mathbf{q}=(a~b~c~d)^{\mathsf{T}}$ which is used to parameterize rotations of a platform belongs to the special unitary group $\mathbb{SU}(2) :=G$. Group operation with an element $\mathbf{q}_0 \in G$, is defined as the quaternion multiplication $\mathbf{q}_0*\mathbf{q}$, which satisfies conditions of closure, i.e.,  $\mathbf{q}_0,\mathbf{q }\in G \implies \mathbf{q}_0*\mathbf{q} \in G$, and conditions of associativity, i.e., $(\mathbf{q}_0*\mathbf{q})*\mathbf{q} = \mathbf{q}_0* (\mathbf{q}*\mathbf{q})$.  An identity element $\mathbf{e}=(1~0~0~0)^{\mathsf{T}}\in G$ and an inverse element $\mathbf{q}^{-1} =(a~-b~-c~-d)^{\mathsf{T}}\in G$ can be identified such that $\mathbf{q}^{-1}*\mathbf{q}=\mathbf{e}$. Specifically this is identified as the left group operation. Similarly, one can define the group operation for a element $\mathbf{q} \in G$ when operated by $\mathbf{q}_0\in G$ as $\mathbf{q}*\mathbf{q}_0$. This is termed the right group operation. 

\begin{table*}[!t]
\centering
% ensure that we have normalsize text
\normalsize
%\setcounter{equation}{2}
%\begin{tabularx}{\linewidth}{c c}
\begin{tabular}{ |c|}
%\label{eqn4:filt_summary}
        %\scriptsize
        %\boxed{
        \hline
        {$\begin{array}{c}
	%\mathrm{\textbf{Filter Equations}}\\
        \begin{array}{c c c c}
        \mathrm{\emph{Observer model}}  &\mathrm{\emph{Measurement model}} &\mathrm{\emph{State Error}} &\mathrm{\emph{Output Error}}\\
%        \multicolumn{4}{l}{\mathrm{EKF:}}\\
%            \begin{pmatrix}
%                \dot{\hat{\mathbf{q}}}\\
%                \dot{\hat{\omega}}_b
%            \end{pmatrix}
%            
%       
%                =
%            \begin{pmatrix}
%               \frac{1}{2}\hat{\mathbf{q}}*(\boldsymbol{\omega}_m-\hat{\boldsymbol{\omega}}_b )\\
%               \mathbf{0}
%            \end{pmatrix}
%             +
%            K\mathcal{E}
%            &\begin{pmatrix}
%                \hat{\mathbf{y}}_a\\
%                \hat{\mathbf{y}}_b\\
%            \end{pmatrix}
%                =
%            \begin{pmatrix}
%               -\hat{\mathbf{R}}_{\mathbf{q}}^\mathsf{T}\mathbf{g}_e\\
%               \hat{\mathbf{R}}_{\mathbf{q}}^\mathsf{T}\mathbf{b}_e
%            \end{pmatrix}
%           &\eta       =~      \begin{pmatrix}
%                            \hat{\mathbf{q}}-\mathbf{q}    \\
%                            \hat{\omega}_b-\omega_b
%                            \end{pmatrix}
%            &\mathcal{E} =~    \begin{pmatrix}
%                             \mathbf{y}_{a}-\mathbf{\hat{y}_a}\\
%                             \mathbf{y}_{b}-\mathbf{\hat{y}_b}\\
%                             \end{pmatrix}\\
        \multicolumn{4}{l}{\mathrm{LIEKF:}}\\
            \begin{pmatrix}
                \dot{\hat{\mathbf{q}}}\\
                \dot{\hat{\omega}}_b
            \end{pmatrix}
                =
            \begin{pmatrix}
               \frac{1}{2}\hat{\mathbf{q}}*(\boldsymbol{\omega}_m-\hat{\boldsymbol{\omega}}_b )\\
               \mathbf{0}
            \end{pmatrix}
             +
            \begin{pmatrix}
               \hat{\mathbf{q}}*L_q\mathcal{E}\\
               L_{\omega_b}\mathcal{E}\\
            \end{pmatrix}\
            &\begin{pmatrix}
                \hat{\mathbf{y}}_a\\
                \hat{\mathbf{y}}_b\\
            \end{pmatrix}
                =
            \begin{pmatrix}
               -\hat{\mathbf{R}}_{q}^\mathsf{T}\mathbf{g}_e\\
               \hat{\mathbf{R}}_{q}^\mathsf{T}\mathbf{b}_e
            \end{pmatrix}
           &\pmb{\eta}       =~      \begin{pmatrix}
                            \mathbf{q}^{-1}*\hat{\mathbf{q}}\\
                            \hat{\omega}_b-\omega_b
                            \end{pmatrix}
            &\mathcal{E} =~    \begin{pmatrix}
                             -\hat{\mathbf{R}}_q^\mathsf{T}\mathbf{g}_e-\mathbf{y}_a\\
                             \hat{\mathbf{R}}_q^\mathsf{T}\mathbf{b}_e-\mathbf{y}_b\\
                             \end{pmatrix}\\
        \multicolumn{4}{l}{\mathrm{RIEKF:}}\\
            \begin{pmatrix}
                \dot{\hat{\mathbf{q}}}\\
                \dot{\hat{\omega}}_b
            \end{pmatrix}
                =
            \begin{pmatrix}
               \frac{1}{2}\hat{\mathbf{q}}*(\boldsymbol{\omega}_m-\hat{\boldsymbol{\omega}}_b )\\
               \mathbf{0}
            \end{pmatrix}
             +
            \begin{pmatrix}
               L_q\mathcal{E}*\hat{\mathbf{q}}\\
               L_{\omega_b}\mathcal{E}\\
            \end{pmatrix}\
            &\begin{pmatrix}
                \hat{\mathbf{y}}_a\\
                \hat{\mathbf{y}}_b\\
            \end{pmatrix}
                =
            \begin{pmatrix}
               -\hat{\mathbf{R}}_{q}^\mathsf{T}\mathbf{g}_e\\
               \hat{\mathbf{R}}_{q}^\mathsf{T}\mathbf{b}_e
            \end{pmatrix}
           &\pmb{\eta}       =~      \begin{pmatrix}
                            \hat{\mathbf{q}}*\mathbf{q}^{-1}\\
                            \mathbf{R}_q(\hat{\omega}_b-\omega_b)
                            \end{pmatrix}
            &\mathcal{E} =~    \begin{pmatrix}
                             -\mathbf{g}_{e}-\hat{\mathbf{R}}_q\mathbf{y}_a\\
                             \mathbf{b}_{e}-\hat{\mathbf{R}}_q\mathbf{y}_b\\
                             \end{pmatrix}\\
                             \end{array}\end{array}$}\\
                             \\
                             \hline
%       \end{array}\\
%       \end{array}\\
%        }
%\end{equation}}
\end{tabular}

\caption{The main filtering equations related to the RIEKF, and the LIEKF. Derivation of each filter is detailed in \cite{Martin2009} .}\label{tbl:generic_filters}
\end{table*}

  \vspace{0.2cm}
\noindent
 \emph{\textbf{Definition 2}}: A smooth map $\phi_g$ is a \emph{Group action} on a set (more generally a manifold) $\mathcal{M}$, if  $(g,m)\in G \times \mathcal{M} \Rightarrow \phi_g(m)\in \mathcal{M} $, s.t.  $ \phi_e(m)=m$ and $\phi_{g_1}(\phi_{g_2}(m))=\phi_{g_1\diamond g_2}(m)$.

 %As an example, the group operation $\phi_{g_1}(g_2)\equiv g_1 \diamond g_2\in G $ can be considered as a trivial smooth map, where a group $G$ acts on itself. More examples of group actions are given by the equation set (\ref{eqn:g_actions}).

 %example - velocity transformation
As an example take linear velocity $\mathbf{v}\in \mathbb{R}^3$ expressed in frame $\{B\}$, and consider the operation of expressing $\mathbf{v}$ in a reference frame $\{E\}$. This mapping is defined as $\phi_{g}(\mathbf{v})=\mathbf{R}_{q}\mathbf{v}\in \mathbb{R}^3$ where the rotation matrix $\mathbf{R}_q=g\in \mathbb{SO}(3)$ is used to specify rotations from frame $\{E\}$ to frame $\{B\}$ . Consider the left group operation defined by matrix multiplication and group elements $g_1,g_2,e \in \mathbb{SO}(3)$ where $g_1=\mathbf{R}_{q_1}$, $g_2=\mathbf{R}_{q_2}$, and $e=\mathbf{I}_3 $ (Identity matrix).  The smooth map $\phi_{g}$ can easily be identified as a group action according to \textit{Definition 2} since $\phi_e(\mathbf{v})=\mathbf{v}$, and $\phi_{g_1}(\phi_{g_2}(\mathbf{v}))= \mathbf{R}_{q_1}\mathbf{R}_{q_2}\mathbf{v} = \phi_{g_1\diamond g_2}(\mathbf{v})$.
 % As an example velocity transformation between two frames can be considered as a group action acting on the velocity space $V=R^3\times TSO(3)$. A group element in G acts on the velocity space V to give $\phi_g(v)=()\in V$, where  $ \phi_e(a)=a$ and  $\phi_{g_1}(\phi_{g_2}(a))=\phi_{g_1\diamond g_2}(a)$ are trivial results for velocity transformation between frames.
 
 \vspace{0.2cm}
\noindent
\emph{\textbf{Definition 3}}: The system $\dot{\mathbf{x}}=f(\mathbf{x},\mathbf{u})$ is \emph{G-invariant} if $f(\varphi_g(\mathbf{x}),\psi(\mathbf{u}))=D\varphi_g(\mathbf{x})\circ f(\mathbf{x},\mathbf{u})$, for group actions $\varphi_g(\mathbf{x})$ on states $\mathbf{x}\in X$, and $\psi_g(\mathbf{u})$  on inputs $\mathbf{u}\in U$.
%Definition 3 is as defined in [], and is not restated here since its not explicitly used in the work of this paper.

 %As an example, see (\ref{eqn:g_invar}). The system model (\ref{eqn:sys_model}) is found to be \emph{G-invariant} for the set of smooth maps given by (\ref{eqn:g_actions}). This implies that the system dynamics is invariant (remains unchanged) under the group actions. A symmetry is an operation that one can perform on a system (i.e. smooth maps) which results in certain features (i.e. invariants) of the system to remain unchanged. Therefore, the group operations (\ref{eqn:g_actions}) can be identified as the symmetries of the system.

 %\noindent An example of a \emph{G-invariant} system for a given set of smooth maps is explained by equation \ref{eqn:g_invar}.
   \vspace{0.2cm}
\noindent
 \emph{\textbf{Definition 4}}: The output $y=h(\mathbf{x},\mathbf{u})$ is \emph{G-equivariant} if $h(\varphi_g(\mathbf{x}),\psi_g(\mathbf{u}))=\rho_g(\mathbf{y})$, for group action $\rho_g(\mathbf{y})$ on outputs $\mathbf{y}\in Y$.

 In \emph{Definition }3, operator $D$  denotes time differentiation and operator $\circ$ denotes function composition. \emph{Definitions }3 and 4 state the necessary conditions for a given system and measurement model to be considered as invariant, when subjected to smooth maps $\varphi_g,~\psi_g,~\rho_g$.  

 %As an example, see (\ref{eqn:g_equivar}). The measurement model (\ref{eqn:meas model}) is found to be \emph{G-equivariant} for the set of smooth maps given by (\ref{eqn:g_actions}).

 %example - \noindent An example of a G-equivariant output is explained by equation ().
  \vspace{0.2cm}
  \noindent
\emph{\textbf{Definition 5}}: A function $\lambda(X,Y,U)$ is an \emph{Invariant} quantity if $\phi_g(\lambda):=\lambda(\varphi_g(X), \rho_g(Y), \psi_g(U)) = \lambda(X,Y,U)$.

%For clarity of presentation, further clarifications and examples related to each of these definitions are introduced as they are used in the design process of the filter.
 As an example, consider the quantity $\lambda=\mathbf{R}_{q}^\mathsf{T}\boldsymbol{\omega}$. This quantity remains unchanged under the group actions $\varphi_{g_0}(\mathbf{R}_{q})=\mathbf{R}_{q_0}\mathbf{R}_{q}$ and  $\psi_{g_0}(\boldsymbol{\omega})= \mathbf{R}_{q_0}\boldsymbol{\omega}$ for group element $g_0=\mathbf{R}_{q_0}\in\mathbb{ SO}(3)$. i.e., $\phi_g(\lambda)~=(\mathbf{R}_{q_0}\mathbf{R}_{q})^\mathsf{T}(\mathbf{R}_{q_0}\boldsymbol{\omega})~=\mathbf{R}_{q}^\mathsf{T}\boldsymbol{\omega}~=\lambda$.

%\subsection{Symmetry preserving design approach}

The process of deriving an IEKF follows four main steps.
\begin{enumerate}
\item Find the symmetries relevant to given system and measurement models. A symmetry is an operation that can be performed on a system (i.e. smooth maps) which results in certain features (i.e. invariants) of the system to remain unchanged. Therefore, the symmetries of a system can be found  by identifying the smooth maps which satisfy \textit{Definitions} 3 and 4 for the models given in equations (\ref{eqn4:qsys_model}) and (\ref{eqn4:qmeas_model}). 
\item Identify invariants of the system which satisfy \textit{Definition} 5. The procedure allows to identify invariant definitions for state error and output error that are more geometrically consistent than the traditional linear state error and output error definitions used in an EKF.
\item Formulate the symmetry preserving observer as proposed in \textit{Theorem 1} in \cite{Bonnabel2008}. Identify the error state dynamics related to the observer, using the invariant quantities found in the previous step. 
\item Stabilize the error state dynamics of the filter using a method such as Lyapunov based design or an error state Kalman filtering approach. When an error state Kalman filtering approach is used the resulting filter is termed as the IEKF. 
\end{enumerate}
The standard IEKF design process for an attitude estimation problem is presented in detail in the work of \cite{Martin2009}, hence not repeated here. This paper specifically proposes modifications relevant to step 2 of this process where a new invariant error definition is introduced; and step 4 of the process where a stability analysis guarantees local asymptotic convergence of the filter proposed in this work. These steps are discussed in detail while the rest of the filter design procedure is summarized. Readers are directed to \cite{Martin2009} for a detailed treatment of the summarized steps.

\section{Methodology}

\begin{figure}[h!]
  \centering
  \includegraphics[width=3.4 in]{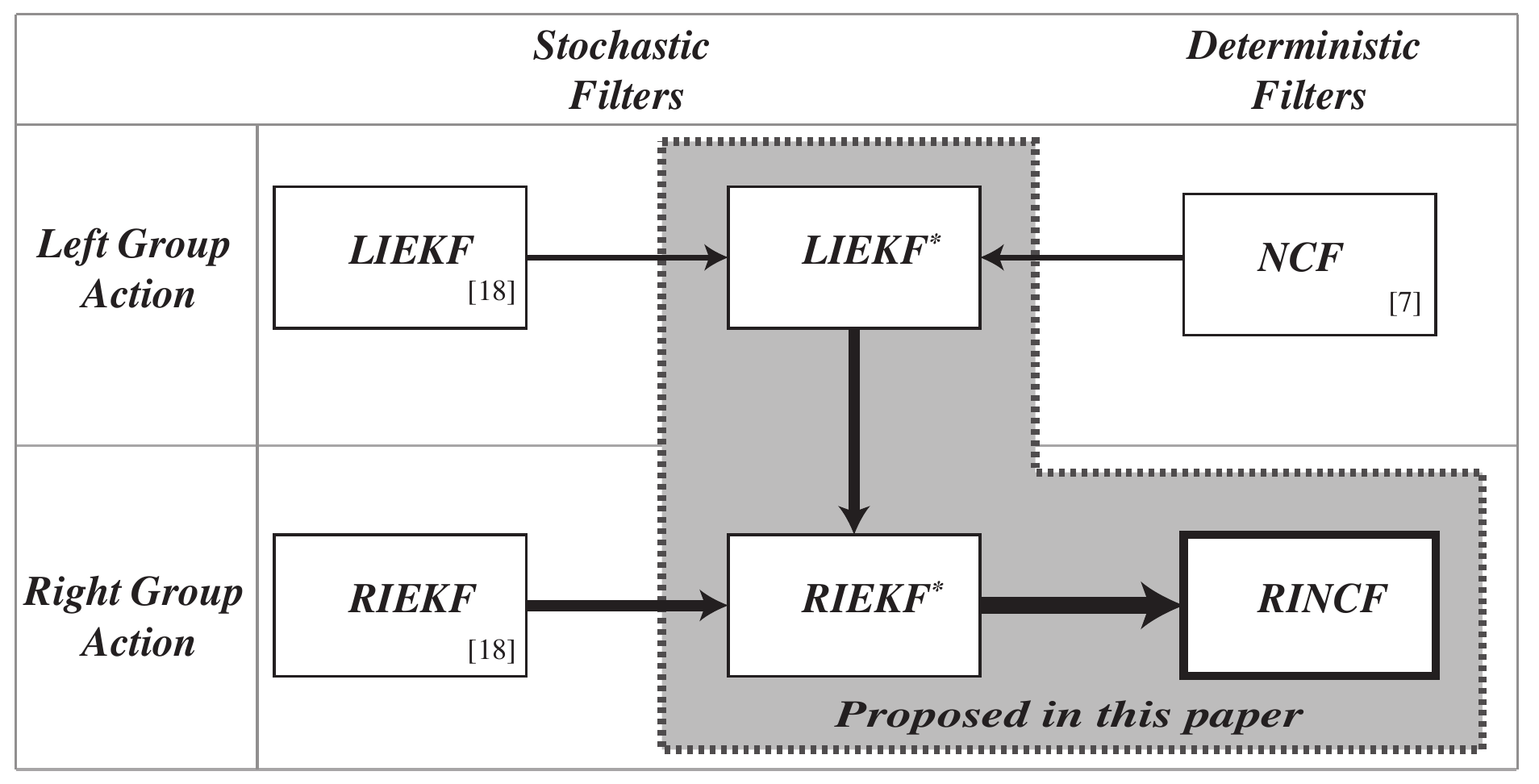}\\
  \caption{The overview of this study and filters designed as part of this work.}\label{fig:paper_flow}
\end{figure}

Fig.\ref{fig:paper_flow} summarizes the development process of the RINCF proposed in this paper. The objective of this study is to find a stochastic steady state equivalent of the NCF, i.e., a stochastic filter having a structure similar to NCF and having a converging  set of constant steady state gains. The resulting filter should have gains which are related to the system noise variables and should be able to perform as a computationally efficient deterministic filter by exploiting its steady state convergence behaviour. 
%Moreover with the expected steady state convergence behaviour, the filter will be able to perform as a deterministic filter with minimal computational overhead.

 The investigation carried out in this paper starts by analyzing well established nonlinear attitude filters available in literature, i.e., the Left IEKF (\textbf{LIEKF}), the Right IEKF (\textbf{RIEKF}) and the NCF. Then this paper identifies modifications to IEKFs' output error definitions which allow them to act as stochastic equivalents for the NCF. These modifications result in LIEKF$^*$ and its symmetric right group equivalent RIEKF$^*$. Next the paper investigates the gain convergence behaviour of these two versions of filters and identifies that RIEKF$^*$ has the best convergence behaviour. Finally, the paper formulates the deterministic form of RIEKF$^*$. This novel formulation is defined as the RINCF.  The RINCF is  the stochastic steady state equivalent form for the right group version of the NCF. 
 
 %The $RINCF$ exhibits easy system noise based gain tuning capability and implementation with minimal computational overhead. 
% include figure

% The following section presents the final filtering equations realized using the IEKF design approach for an orientation estimation problem. This work specifically introduces the following two steps, a) identify a candidate steady state deterministic form, and b) find optimal steady state gains for the filter, in order to realize the RINCF proposed in this work. 

%\subsection{EKF, the IEKF and the NCF}

\subsection{Relationship between the $NCF$ and the $IEKF$} 

An  IEKF design can be followed to realize a filter for the model given by equations (\ref{eqn4:qsys_model}) and (\ref{eqn4:qmeas_model}), as described in work presented in \cite{Martin2009}. It follows the symmetry preserving observer design procedure to derive two forms of the filter: the LIEKF and the RIEKF. The difference between these two filters is that LIEKF considers the left group action  and RIEKF considers the right group action relevant to the rotation group; i.e., for group elements $g=\mathbf{q},g_0=\mathbf{q}_0\in \mathbb{SU}(2)$,   $g_0\diamondsuit g=\mathbf{q}_0*\mathbf{q}$ and $g_0\diamondsuit g= \mathbf{q}*\mathbf{q}_0$ denotes the left and right group actions respectively. These filters have adaptive gains which are based on system noise variables and only implies local convergence due to the use of linearized error state models.
 The resulting filtering equations of LIEKF and RIEKF for the AHRS problem are summarized in Table \ref{tbl:generic_filters}.  The symbol $\wedge$ is used to identify the estimate of a variable. The scalars $L_q,~L_{\omega_b}$ are the gains of the filter for states $\pmb{q} $ and  $\pmb{\omega}_b$ respectively. The variable $\pmb{\eta}$ denotes the state error and $\mathcal{E}$ denotes the output error. The observer model, state error and output error in Table \ref{tbl:generic_filters} are designed such that they are invariant as defined in \textit{Definition 5}.

%\begin{figure*}[t!]
 % \centering
 % \includegraphics[width=6.9 in]{ahrs_gains1}\\
 % \caption{Gain matrix $K(t)$ of (a) the EKF, (b) the LIEKF,  and (c) the RIEKF when following a smooth trajectory}\label{fig:ahrs_gains1}
%\end{figure*}

\begin{table}
\centering
\begin{tabular}{|c|}    
\hline
    {$
    %{\begin{equation*}
        %\scriptsize
       % \boxed{
        \begin{array}{l}
         ~~~~~~~~~~~~~~~~~~~~~~~~~~~~~~~~~~NCF\\
        \mathrm{Initialization}: ~~~~\begin{array}{c l}
                    \hat{\mathbf{x}} &=~\begin{bmatrix}1 &\mathbf{0} &\mathbf{0} \end{bmatrix}^\mathsf{T}\\
                    \end{array}\\
        \mathrm{Filter~Equations}:\\
        \begin{array}{c}
        \begin{array}{c c}
            \begin{pmatrix}
                \dot{\hat{\mathbf{q}}}\\
                \dot{\hat{\omega}}_b
            \end{pmatrix}
                =
            \begin{pmatrix}
               \frac{1}{2}\hat{\mathbf{q}}*(\boldsymbol{\omega}_m-\hat{\boldsymbol{\omega}}_b )\\
               \mathbf{0}
            \end{pmatrix}
             +
             \begin{pmatrix}
               \hat{\mathbf{q}}*k_p\mathcal{E}\\
               k_i\mathcal{E}\\
            \end{pmatrix}\\
            
            \begin{pmatrix}
                \hat{\mathbf{y}}_a\\
                \hat{\mathbf{y}}_b\\
            \end{pmatrix}
                =
            \begin{pmatrix}
               -\hat{\mathbf{R}}_{\mathbf{q}}^\mathsf{T}g_e\\
               \hat{\mathbf{R}}_{\mathbf{q}}^\mathsf{T}b_e
            \end{pmatrix}\\
            %\mathrm{\emph{Output Error}}~~~~~~~~
            \mathcal{E} =~
                             k_1(\mathbf{\hat{y}_a}\times\mathbf{y}_{a})+
                             k_2(\mathbf{\hat{y}_b}\times\mathbf{y}_{b})\\
         \end{array}\\
        \end{array}\\
         \begin{array}{l}
         \mathrm{Gain~selection}:~~ \begin{bmatrix}
                                        k_p& k_i& k_1 &k_2       \end{bmatrix}^\mathsf{T} >0\\
        \end{array}\\
        \end{array}
        $}\\
   % \end{equation*}}\\
   \hline
\end{tabular}
\caption{The nonlinear complementary filter for attitude estimation}\label{tbl:NCF}
\end{table}    
 
The NCF for the attitude estimation problem is summarized in Table \ref{tbl:NCF}, where $k_p, k_d, k_1$ and $k_2$ denote the gains related to the filter. NCF exhibits good convergence and wide acceptance as a deterministic, low cost form, applicable for attitude filtering. The strength of the NCF is that it uses a constant set of gains to achieve stable response  and has global stability implications as reported in \cite{Mahony2008}. %However, the gains of the NCF are hand picked by trial and error, which is counter intuitive to system noise based tuning capability available with stochastic filters.
 
% Fig.\ref{fig:ahrs_gains1} illustrates the gain stabilization behaviour of the  EKF, LIEKF, and RIEKF  when a platform is following a smooth trajectory. All filters converge to the correct orientation estimate during operation. Interestingly, the RIEKF exhibits convergence of its dominant gains to a constant set of gains. This is an important characteristic since it implies that the filter converges to a constant gain filter like the NCF during operation. Furthermore, unlike the NCF these constant gains are base on the system noise variables. Hence this paper investigates whether theres a connection between the NCF and the IEKFs for attitude filtering; and if this relationship can be exploited to device an effective tuning scheme for the NCF.

  It can be clearly seen that the NCF given in Table \ref{tbl:NCF}  is directly comparable to the LIEKF given in Table \ref{tbl:generic_filters}, with the exception of their output error definitions $\mathcal{E}$. The NCF takes a cross product form for its output error definition while the LIEKF takes a quaternion error form. In order to realize an equivalent stochastic form for the NCF, the output error definition $\mathcal{E}$ of the LIEKF in Table \ref{tbl:generic_filters} should be modified as follows.
      \begin{equation}\label{eqn:operror}
     \mathcal{E} =~      \begin{pmatrix}
                  \mathbf{\mathbf{y}_a}\times\hat{\mathbf{y}}_{a}\\
                  \mathbf{\mathbf{y}_b}\times\hat{\mathbf{y}}_{b}\\
                  \end{pmatrix}
     \end{equation}
\noindent
The sign change of the error definition is common in Luenberger type observers, which cancels the effect by producing gains with the corresponding sign change. The next section proceeds to apply the cross product modification given in equation (\ref{eqn:operror}) in the invariant filter design process.

 \subsection{ The $LIEKF*$ and the $RIEKF*$}

\begin{table*}[t!]
\centering
\caption{Summary of LIEKF$^*$ and RIEKF$^*$}\label{tbl:ch4_LIEKFmod_summary}
 \begin{tabular}{|c| c|}
 \hline
 {$\begin{array}{l}
 ~~~~~~~~~~~~~~~~~~~~~~~~~~~~~~~~~~~~~~~~~~LIEKF^*\\
        \mathrm{Initialization:}\\
            ~~~~\begin{array}{c c c c}
            \hat{\mathbf{x}} =~\begin{bmatrix}1 &\mathbf{0} &\mathbf{0} \end{bmatrix}^\mathsf{T},
            &P= \mathbf{I},
            &Q                =~ \mathrm{E}(\mathbf{w}\mathbf{w}^\mathsf{T}),
            &R                =~\mathrm{E}(\boldsymbol{\nu}\boldsymbol{\nu}^\mathsf{T})\\
            \end{array}\\
        \mathrm{Filter~Equations:}\\
        \begin{array}{c}
        \begin{array}{c }
            \begin{pmatrix}
                \dot{\hat{\mathbf{q}}}\\
                \dot{\hat{\omega}}_b
            \end{pmatrix}
                =
           \begin{pmatrix}
               \frac{1}{2}\hat{\mathbf{q}}*(\boldsymbol{\omega}_m-\hat{\boldsymbol{\omega}}_b )\\
               \mathbf{0}
            \end{pmatrix}
             +
            \begin{pmatrix}
               \hat{\mathbf{q}}*L_q\mathcal{E}\\
               L_{\omega_b}\mathcal{E}\\
            \end{pmatrix}\\
             \begin{pmatrix}
                \hat{\mathbf{y}}_a\\
                \hat{\mathbf{y}}_b\\
            \end{pmatrix}
                =
             \begin{pmatrix}
               -\hat{\mathbf{R}}_{\mathbf{q}}^\mathsf{T}\mathbf{g}_e\\
               \hat{\mathbf{R}}_{\mathbf{q}}^\mathsf{T}\mathbf{b}_e
             \end{pmatrix} 
             ~~~~~~~\mathcal{E} =~    \begin{pmatrix}
                             \mathbf{\hat{y}_a}\times\mathbf{y}_{a}\\
                             \mathbf{\hat{y}_b}\times\mathbf{y}_{b}\\
                             \end{pmatrix}\\
       %     & \mathrm{\emph{State Prediction}} & \mathrm{\emph{State Correction}} \\%& \mathrm{\emph{Output Prediction}}\\
        \end{array}\\
        \begin{array}{c}
            %\mathrm{\emph{Output Error}}~~
%            \mathcal{E} =~    \begin{pmatrix}
%                             \hat{\mathbf{R}}_\mathbf{q}(\mathbf{\hat{y}_a}\times\mathbf{y}_{a})\\
%                             \hat{\mathbf{R}}_\mathbf{q}(\mathbf{\hat{y}_b}\times\mathbf{y}_{b})\\
%                             \end{pmatrix}\\
            %&\mathrm{\emph{Gain Matrix}}~~K       =\begin{bmatrix}
             %               -L_q &-L_{\omega_b}
              %              \end{bmatrix}^\mathsf{T}
             % \mathrm{\emph{Gain Computation}}\\
             K=\begin{bmatrix}
             -L_q &-L_{\omega_b}
              \end{bmatrix}^\mathsf{T}
         \end{array}\\
            \end{array}\\
        \mathrm{Filtering~Matrices(used~to~compute~Kalman~gain~K):}\\
            ~
             \begin{array}{l l}
            A=~\begin{bmatrix}
                [\hat{\omega}_b - \omega_m]_{\times}  &-\frac{1}{2}\mathbf{I}\\
                \mathbf{0}   &\mathbf{0}
             \end{bmatrix}
            &C=~\begin{bmatrix}
                2[\hat{\mathbf{R}}_{\mathbf{q}}^\mathsf{T}\mathbf{g}_e]_{\times}^2  &\mathbf{0}\\
                2[\hat{\mathbf{R}}_{\mathbf{q}}^\mathsf{T}\mathbf{b}_e]_{\times}^2  &\mathbf{0}
             \end{bmatrix}\\
             M=~\begin{bmatrix}
                0.5\mathbf{I}  &\mathbf{0}\\
                \mathbf{0}  &-\mathbf{I}
             \end{bmatrix}
             &N=~\begin{bmatrix}
                \mathbf{I}+[\hat{\mathbf{R}}_{\mathbf{q}}^\mathsf{T}\mathbf{g}_e]_{\times}  &\mathbf{0}\\
                \mathbf{0}               &\mathbf{I}-[\hat{\mathbf{R}}_{\mathbf{q}}^\mathsf{T}\mathbf{b}_e]_{\times}
             \end{bmatrix}\\
            \end{array}\\
            \end{array}
    $} 
    &
    {$\begin{array}{l}
 ~~~~~~~~~~~~~~~~~~~~~~~~~~~~~~~~~~~~~~~~~~RIEKF^*\\
        \mathrm{Initialization:}\\
            ~~~~\begin{array}{c c c c}
            \hat{\mathbf{x}} =~\begin{bmatrix}1 &\mathbf{0} &\mathbf{0} \end{bmatrix}^\mathsf{T},
            &P= \mathbf{I},
            &Q                =~ \mathrm{E}(\mathbf{w}\mathbf{w}^\mathsf{T}),
            &R                =~\mathrm{E}(\boldsymbol{\nu}\boldsymbol{\nu}^\mathsf{T})\\
            \end{array}\\
        \mathrm{Filter~Equations:}\\
        \begin{array}{c}
        \begin{array}{c }
            \begin{pmatrix}
                \dot{\hat{\mathbf{q}}}\\
                \dot{\hat{\omega}}_b
            \end{pmatrix}
                =
           \begin{pmatrix}
               \frac{1}{2}\hat{\mathbf{q}}*(\boldsymbol{\omega}_m-\hat{\boldsymbol{\omega}}_b )\\
               \mathbf{0}
            \end{pmatrix}
             +
            \begin{pmatrix}
               L_q\mathcal{E}*\hat{\mathbf{q}}\\
               \hat{\mathbf{R}}^{\mathsf{T}}_\mathbf{q}L_{\omega_b}\mathcal{E}\\
               %\hat{\mathbf{q}}^{-1}*L_{\omega_b}\mathcal{E}*\hat{\mathbf{q}}\\
            \end{pmatrix}\\
             \begin{pmatrix}
                \hat{\mathbf{y}}_a\\
                \hat{\mathbf{y}}_b\\
            \end{pmatrix}
                =
             \begin{pmatrix}
               -\hat{\mathbf{R}}_{\mathbf{q}}^\mathsf{T}\mathbf{g}_e\\
               \hat{\mathbf{R}}_{\mathbf{q}}^\mathsf{T}\mathbf{b}_e
             \end{pmatrix} 
             ~~~~~~~\mathcal{E} =~    \begin{pmatrix}
                             \hat{\mathbf{R}}_\mathbf{q}(\mathbf{\hat{y}_a}\times\mathbf{y}_{a})\\
                             \hat{\mathbf{R}}_\mathbf{q}(\mathbf{\hat{y}_b}\times\mathbf{y}_{b})\\
                             \end{pmatrix}\\
       %     & \mathrm{\emph{State Prediction}} & \mathrm{\emph{State Correction}} \\%& \mathrm{\emph{Output Prediction}}\\
        \end{array}\\
        \begin{array}{c}
            %\mathrm{\emph{Output Error}}~~
%            \mathcal{E} =~    \begin{pmatrix}
%                             \hat{\mathbf{R}}_\mathbf{q}(\mathbf{\hat{y}_a}\times\mathbf{y}_{a})\\
%                             \hat{\mathbf{R}}_\mathbf{q}(\mathbf{\hat{y}_b}\times\mathbf{y}_{b})\\
%                             \end{pmatrix}\\
            %&\mathrm{\emph{Gain Matrix}}~~K       =\begin{bmatrix}
             %               -L_q &-L_{\omega_b}
              %              \end{bmatrix}^\mathsf{T}
             % \mathrm{\emph{Gain Computation}}\\
             K=\begin{bmatrix}
             -L_q &-L_{\omega_b}
              \end{bmatrix}^\mathsf{T}
         \end{array}\\
            \end{array}\\
        \mathrm{Filtering~Matrices(used~to~compute~Kalman~gain~K):}\\
            ~
             \begin{array}{l l}
            A=~\begin{bmatrix}
                \mathbf{0}   &-\frac{1}{2}\mathbf{I}\\
                \mathbf{0}   &\hat{\mathbf{R}}_{\mathbf{q}}(\omega_m-\hat{\omega}_b)
             \end{bmatrix}
            &C=~\begin{bmatrix}
                2[\mathbf{g}_e]_{\times}^2  &\mathbf{0}\\
                2[\mathbf{b}_e]_{\times}^2  &\mathbf{0}
             \end{bmatrix}\\
             M=~\begin{bmatrix}
                0.5\mathbf{I}  &\mathbf{0}\\
                \mathbf{0}  &-\mathbf{I}
             \end{bmatrix}
             &N=~\begin{bmatrix}
                \mathbf{I}+[\mathbf{g}_e]_{\times}  &\mathbf{0}\\
                \mathbf{0}               &\mathbf{I}-[\mathbf{b}_e]_{\times}
             \end{bmatrix}\\
            \end{array}\\
            \end{array}
    $} 
     \\
 \hline
 \end{tabular}
 \end{table*}

To be congruent with the symmetry preserving framework, the output error definition given by (\ref{eqn:operror}) should satisfy conditions of invariance. The relevant symmetries of the LIEKF when operated by a group element $(\mathbf{q}_0,~\boldsymbol{\omega}_{b0})\in G = \mathbb{SU}(2)\times \mathbb{R}^3$, can be found by applying \textit{Definitions} 3 and 4 as follows.
\begin{equation*}
    \begin{array}{c c c}
          \varphi_{g}
            \begin{pmatrix}
                \mathbf{q}\\
                \mathbf{\omega}_b\\

            \end{pmatrix}
                =
             \begin{pmatrix}
                \mathbf{q}_0*\mathbf{q}\\
                \mathbf{\omega}_b + \mathbf{\omega}_{b0}\\
              \end{pmatrix},

            &\psi_g \begin{pmatrix}
                \mathbf{\omega}_m\\
                \mathbf{g}_e\\
                \mathbf{b}_e\\
             \end{pmatrix}
                =
             \begin{pmatrix}
                \mathbf{\omega}_m + \mathbf{\omega}_{b0}\\
                \mathbf{R}_{q_0}\mathbf{g}_e\\
                \mathbf{R}_{q_0}\mathbf{b}_e
              \end{pmatrix},
            \\
            \rho_g \begin{pmatrix}
                \mathbf{y}_a\\
                \mathbf{y}_b
             \end{pmatrix}
                =
             \begin{pmatrix}
               \mathbf{y}_a\\
               \mathbf{y}_b
              \end{pmatrix}
    \end{array}
\end{equation*}
\noindent
Using these mappings we can show that the cross product error term (\ref{eqn:operror}) is invariant according to \textit{Definition} 5.
    \begin{equation*}
    \begin{array}{c}
     \phi_g(\mathcal{E}(\hat{x},u,y)) =\mathcal{E}(\varphi_g(\hat{x}),\psi_g(u),\rho_g(y)) \\
          \phi_g\begin{pmatrix}
                  \mathbf{\mathbf{y}_a}\times-\hat{\mathbf{R}}_{q}^\mathsf{T}\mathbf{g}_e\\
                  \mathbf{\mathbf{y}_b}\times\hat{\mathbf{R}}_{q}^\mathsf{T}\mathbf{b}_e\\
                  \end{pmatrix} =
       \begin{pmatrix}
                  \mathbf{\mathbf{y}_a}\times-\hat{\mathbf{R}}_{q}^\mathsf{T}\mathbf{R}_{q0}^\mathsf{T}\mathbf{R}_{q0}\mathbf{g}_e\\
                  \mathbf{\mathbf{y}_b}\times\hat{\mathbf{R}}_{q}^\mathsf{T}\mathbf{R}_{q0}^\mathsf{T}\mathbf{R}_{q0}\mathbf{b}_e\\
       \end{pmatrix}  \\
       =\begin{pmatrix}
                  \mathbf{\mathbf{y}_a}\times-\hat{\mathbf{R}}_{q}^\mathsf{T}\mathbf{g}_e\\
                  \mathbf{\mathbf{y}_b}\times\hat{\mathbf{R}}_{q}^\mathsf{T}\mathbf{b}_e\\
                  \end{pmatrix} = \mathcal{E}(\hat{x},u,y)
     \end{array}
    \end{equation*}
    
 \noindent
The modified LIEKF which incorporates the cross product error given by  (\ref{eqn:operror}) is derived by following the IEKF design process. This modified LIEKF is denoted as LIEKF$^*$.

By following a similar procedure, one can find the modified RIEKF which uses the cross product error term in its formulation.  The cross product error term applicable for right group actions is given in equation (\ref{eqn:operror2}).  
    \begin{equation}\label{eqn:operror2}
     \mathcal{E} =~      \begin{pmatrix}
                  \hat{\mathbf{R}}_q(\mathbf{\mathbf{y}_a}\times\hat{\mathbf{y}}_{a})\\
                  \hat{\mathbf{R}}_q(\mathbf{\mathbf{y}_b}\times\hat{\mathbf{y}}_{b})\\
                  \end{pmatrix}
     \end{equation}
The following set of equations identify the symmetries relevant to the RIEKF. 
 \begin{equation*}
    \begin{array}{c c c}
          \varphi_{g}
            \begin{pmatrix}
                \mathbf{q}\\
                \mathbf{\omega}_b\\

            \end{pmatrix}
                =
             \begin{pmatrix}
                \mathbf{q}*\mathbf{q}_0\\
                \mathbf{R}_{q_0}^\mathsf{T}\mathbf{\omega}_b + \mathbf{\omega}_{b0}\\
              \end{pmatrix},

            &\psi_g \begin{pmatrix}
                \mathbf{\omega}_m\\
                \mathbf{g}_e\\
                \mathbf{b}_e\\
             \end{pmatrix}
                =
             \begin{pmatrix}
                \mathbf{R}_{q_0}^\mathsf{T}\mathbf{\omega}_m + \mathbf{\omega}_{b0}\\
                \mathbf{g}_e\\
                \mathbf{b}_e
              \end{pmatrix},\\
            \rho_g \begin{pmatrix}
                \mathbf{y}_a\\
                \mathbf{y}_b
             \end{pmatrix}
                =
             \begin{pmatrix}
               \mathbf{R}_{q_0}^\mathsf{T}\mathbf{y}_a\\
               \mathbf{R}_{q_0}^\mathsf{T}\mathbf{y}_b
              \end{pmatrix}
    \end{array}
\end{equation*}
\noindent
Using these mappings we can verify the invariance of  (\ref{eqn:operror2}) for symmetries relevant to right group actions.

\begin{equation*}
    \begin{array}{c}
     \phi_g(\mathcal{E}(\hat{x},u,y)) =\mathcal{E}(\varphi_g(\hat{x}),\psi_g(u),\rho_g(y)) \\
          \phi_g\begin{pmatrix}
                  \hat{\mathbf{R}}_{q}(\mathbf{\mathbf{y}_a}\times-\hat{\mathbf{R}}_{q}^\mathsf{T}\mathbf{g}_e)\\
                  \hat{\mathbf{R}}_{q}(\mathbf{\mathbf{y}_b}\times\hat{\mathbf{R}}_{q}^\mathsf{T}\mathbf{b}_e)\\
                  \end{pmatrix} =
       \begin{pmatrix}
                  \hat{\mathbf{R}}_{q}\mathbf{R}_{q_0}(\mathbf{R}_{q_0}^\mathsf{T}\mathbf{\mathbf{y}_a}\times-\mathbf{R}_{q_0}^\mathsf{T}\hat{\mathbf{R}}_{q}^\mathsf{T}\mathbf{g}_e)\\
                  \hat{\mathbf{R}}_{q}\mathbf{R}_{q_0}(\mathbf{R}_{q_0}^\mathsf{T}\mathbf{\mathbf{y}_b}\times\mathbf{R}_{q_0}^\mathsf{T}\hat{\mathbf{R}}_{q}^\mathsf{T}\mathbf{b}_e)\\
       \end{pmatrix} \\ =
       \begin{pmatrix}
                  \hat{\mathbf{R}}_{q}(\mathbf{\mathbf{y}_a}\times-\hat{\mathbf{R}}_{q}^\mathsf{T}\mathbf{g}_e)\\
                  \hat{\mathbf{R}}_{q}(\mathbf{\mathbf{y}_b}\times\hat{\mathbf{R}}_{q}^\mathsf{T}\mathbf{b}_e)\\
                  \end{pmatrix} = \mathcal{E}(\hat{x},u,y)
     \end{array}
    \end{equation*}
\noindent
The modified RIEKF which incorporates the cross product error term given by (\ref{eqn:operror2}) is denoted as RIEKF$^*$. The resulting AHRS designs LIEKF$^*$ and RIEKF$^*$, are summarized in Table \ref{tbl:ch4_LIEKFmod_summary}. Here the state covariance matrix is denoted by $P$, system noise covariance is denoted by $Q$, the measurement noise covariance is denoted by $R$ and the Kalman gain is denoted by $K$. The operator $E(\cdot)$ denotes the expected value computation of a random variable. The Jacobian matrices corresponding to system, measurement, process noise and measurement noise models are denoted by $A,C,M$ and $N$ respectively following standard notation. The identity matrix is denoted using $\pmb{I}$ and the operator $[\cdot]_{\times}$ is used to denote standard skew symmetric matrix transformation of a vector. 

\begin{figure*}[t!]
  \centering
  \includegraphics[width=6.9 in]{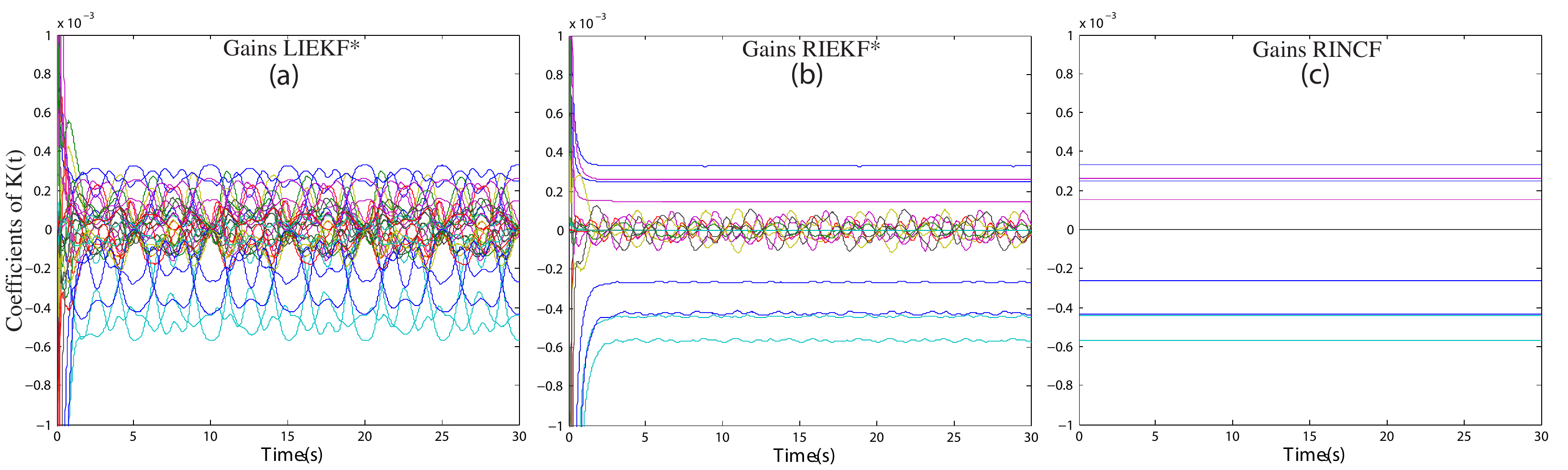}\\
  \caption{Gain matrix $K(t)$ of (a) the RIEKF$^*$, (b) the LIEKF$^*$, and (c) the RINCF for a smooth trajectory}\label{fig:ahrs_gains2}
\end{figure*}

Notice that cross product is a rank deficient operation which causes the measurement noise Jacobian $N$ to lose its rank. Therefore the measurement noise Jacobians are modified as $I+N$ in order to preserve their full rank condition, allowing them to be applied in Kalman gain computation equations.

Fig.\ref{fig:ahrs_gains2} (a) and (b) illustrate the gain stabilization of LIEKF$^*$ and RIEKF$^*$ respectively. It is evident from the graphs that the RIEKF$^*$ has a stabilizing constant set of gains, which is desirable for a fixed gain design. A steady state gain identification for the LIEKF$^*$  will be difficult due to the existence of strong coupling between the gains and the system trajectory.  In the next step we attempt to exploit RIEKF$^*$ further to propose an intuitive gain tuning method for the NCF.

\noindent
\subsection{The RINCF}\label{sec:rincf}
The invariant state error $\boldsymbol{\eta}$ for a right invariant formulation is defined as follows (as given in Table \ref{tbl:generic_filters}-RIEKF).
\begin{equation*}
\pmb{\eta}       =~      \begin{pmatrix}
                            \hat{\mathbf{q}}*\mathbf{q}^{-1}\\
                            \mathbf{R}_q(\hat{\omega}_b-\omega_b)
                            \end{pmatrix}
                         =~    \begin{pmatrix}
                            \boldsymbol{\mu}\\
                           \pmb{\beta}
                            \end{pmatrix}                           
\end{equation*}
Here the rotational error state is denoted by $\boldsymbol{\mu}$ and the gyroscopic bias error state is defined by $\pmb{\beta}$.   The error state dynamic model of RIEKF$^*$ (\ref{es_dynamics_rincf}) can be found by time differentiating  $\pmb{\eta}$  and expressing the resulting model using invariant quantities. 

%es dynamics
\begin{equation}\label{es_dynamics_rincf}
    \begin{array}{c c}
     \begin{pmatrix}
                  \dot{\boldsymbol{\mu}}\\
                 \dot{\boldsymbol{\beta}}
      \end{pmatrix} =
      \begin{pmatrix}
                  -\frac{1}{2}\boldsymbol{\beta}*\boldsymbol{\mu}+L_{\mu}\mathcal{E}\\
                 \boldsymbol{\mu}^{-1}*\hat{I}_{\omega}*\boldsymbol{\mu}\times \boldsymbol{\beta}+\mu^{-1}*L_{\beta}\mathcal{E}*\boldsymbol{\mu}
      \end{pmatrix}
    \\
    \mathcal{E} =
    \begin{pmatrix}
                  \mathbf{e}_g\\
                  \mathbf{e}_b
      \end{pmatrix} =
      \begin{pmatrix}
                 \mathbf{R}_{\mu}\mathbf{g}_e\times \mathbf{g}_e\\
                 \mathbf{R}_{\mu}\mathbf{b}_e\times \mathbf{b}_e\\
                 %\boldsymbol{\mu}*\mathbf{g}_e*\boldsymbol{\mu}^{-1}\times \mathbf{g}_e\\
                 %\boldsymbol{\mu}*\mathbf{b}_e*\boldsymbol{\mu}^{-1}\times \mathbf{b}_e
      \end{pmatrix}
    \end{array}
\end{equation}
\noindent
The invariant quantity $\hat{I}_{\omega}$  is defined as  $\hat{I}_{\omega}=\hat{\mathbf{q}}*(\boldsymbol{\omega}_m-\hat{\boldsymbol{\omega}}_b)*\hat{\mathbf{q}}^{-1}$. The matrix $\mathbf{R}_{\mu}$ denotes the rotation matrix corresponding to the quaternion error state $\pmb{\mu}$. The steady state gain convergence behaviour of RIEKF$^*$ shown in Fig.\ref{fig:ahrs_gains2}, implies that the optimal gain matrix $K$ would mainly consist of a constant set of gains. Consider the following constant gain matrix proposal.

%the gain proposal
%es dynamics
\begin{equation*}
    \begin{array}{c c}
    K=
     \begin{pmatrix}
                  L_{\mu}\\
                  L_{\beta}
      \end{pmatrix} =
      \begin{pmatrix}
                 \mathbf{I}_{\mathbf{a}} &\mathbf{I}_{\mathbf{b}}\\
                 -\mathbf{I}_{\mathbf{c}} &-\mathbf{I}_{\mathbf{d}}
      \end{pmatrix},
      \begin{matrix}
      \mathbf{I}_{\mathbf{a}}=diag([a_1~ a_2~ a_3])\\
      \mathbf{I}_{\mathbf{b}}=diag([b_1~ b_2~ b_3])\\
      \mathbf{I}_{\mathbf{c}}=diag([c_1~ c_2~ c_3])\\
      \mathbf{I}_{\mathbf{d}}=diag([d_1~ d_2~ d_3])
      \end{matrix}
      %
%      \begin{pmatrix}
%                 a_1&   0&      0& 0& 0&    0\\
%                 0&     a_2&    0& 0& b_2&  0\\
%                 0&     0&      0& 0& 0&    b_3\\
%                 c_1&   0&      0& 0& 0&    0\\
%                 0&     c_2&    0& 0& d_2&  0\\
%                 0&     0&      0& 0& 0&    d_3
%      \end{pmatrix}
    \end{array}
\end{equation*}
\noindent
The gain matrix $K$ is substituted in (\ref{es_dynamics_rincf}) and the resulting error state system is linearized using  a small signal assumption, i.e. $\pmb{\mu} \rightarrow (1,~\delta\pmb{\mu}^{\mathsf{T}})^{\mathsf{T}},~ \pmb{\beta}\rightarrow \delta\pmb{\beta} $. The linearized error states corresponding to error states $\boldsymbol{\mu}$ and $\boldsymbol{\beta}$ are denoted as $\delta\boldsymbol{\mu}$ and $\delta\boldsymbol{\beta}$ respectively. Note that $\delta\boldsymbol{\mu}$ is a three dimensional vector since infinitesimal quaternion errors result in the first entry of quaternion error $\boldsymbol{\mu}$ to be $1$. Second order infinitesimal terms of the resulting model are removed to generate the linearized error state model as given in equations  (\ref{eqn:ahrs_es_system}). 
\begin{equation}\label{eqn:ahrs_es_system}
    \begin{array}{c c}
     \begin{pmatrix}
                  \dot{\delta\boldsymbol{\mu}}\\
                 \dot{\delta\boldsymbol{\beta}}
      \end{pmatrix} =
      \begin{pmatrix}
                  -\frac{1}{2}\delta\boldsymbol{\beta}+A_{\mu}\delta\boldsymbol{\mu}\\
                 \hat{I}_{\omega}\times\delta\boldsymbol{\beta}+ A_{\beta}\delta\boldsymbol{\mu}
      \end{pmatrix},
    \\ A_{\mu} =
    \begin{pmatrix}
                  -2g_e^2a_1    &0                      &0\\
                  0             &-2g_e^2a_2-2b_e^2b_2   &0\\
                  0             &0                      &-2b_e^2b_3
      \end{pmatrix},
      \\ A_{\beta} =
    \begin{pmatrix}
                  2g_e^2c_1    &0                      &0\\
                  0             &2g_e^2c_2+2b_e^2d_2   &0\\
                  0             &0                      &2b_e^2d_3
      \end{pmatrix}
    \end{array}
\end{equation}

\noindent
This derivation assumes that the measurement system is calibrated and aligned such that the gravity vector $\mathbf{g}_e=(0,~0,~g_e)^{\mathsf{T}}$ and the magnetic reference $\mathbf{b}_e=(b_e,~ 0,~ 0)^{\mathsf{T}}$. This system is found to be locally asymptotically stable using a candidate Lyapunov function $V$ \cite{Martin2010}.
\begin{equation*}
V~=\delta\boldsymbol{\beta}^\mathsf{T}\delta\boldsymbol{\beta}~ + (2A_{\beta}\delta\boldsymbol{\mu})^{\mathsf{T}}\delta\boldsymbol{\mu} \geq 0
 \end{equation*}
\noindent
The derivative of the Lyapunov function is found to be: % to do - barbellets , lasalles theorums
\begin{equation*}
\begin{array}{c l}
    \dot{V}&=(4A_{\boldsymbol{\beta}}\delta\boldsymbol{\mu})^{\mathsf{T}}A_{\mu}\delta\boldsymbol{\mu}~\\
    &=-4\delta\mu_y^2(2b_2b_e^2 + 2a_2g_e^2)(2d_2b_e^2 + 2c_2g_e^2) \\&- 16b_3b_e^4d_3\delta\mu_z^2 - 16a_1c_1g_e^4\delta\mu_x^2
\end{array}
\end{equation*}
Here $\dot{V}$ is negative semi-definite which implies Lyapunov stability, i.e., $\delta\boldsymbol{\mu}$ and $\delta\boldsymbol{\beta}$ are bounded. $\ddot{V}$ is a polynomial function of $\delta\boldsymbol{\mu}$, $\delta\boldsymbol{\beta}$, $I_{\omega}$ and $\dot{I}_{\omega}$. Hence $\ddot{V}$ is bounded and $\dot{V}$ is uniformly continuous. From Barbalat's lemma $\dot{V}\rightarrow0$ as $t\rightarrow \infty$, hence $\delta\boldsymbol{\mu}\rightarrow0$ as $t\rightarrow \infty$. Analyzing the local error state dynamics (\ref{eqn:ahrs_es_system}), $\dot{\delta\boldsymbol{\mu}}$ is bounded and $\ddot{\delta\boldsymbol{\mu}}$ is a polynomial function of  $\delta\boldsymbol{\mu}$, $\delta\boldsymbol{\beta}$, $I_{\omega}$ and $\dot{I}_{\omega}$. Hence $\ddot{\delta\boldsymbol{\mu}}$ is bounded and  $\dot{\delta\boldsymbol{\mu}}$ is uniformly continuous. From Barbalat's lemma $\dot{\delta\boldsymbol{\mu}}\rightarrow0$ as $t\rightarrow \infty$. Since $\delta\boldsymbol{\mu}\rightarrow0$ as $t\rightarrow \infty$, this means that $\delta{\boldsymbol{\beta}}\rightarrow0$ as $t\rightarrow \infty$. Therefore, the system is locally asymptotically convergent to the equilibrium for any positive parameter selection for $a_1,~a_2,~b_2,~b_3,~c_1,~c_2,~d_2,~d_3$. %Work in \cite{Mahony2008}, carries global stability implications of the filter which is not discussed in this text.

The constant gain $K$ can be found by applying DARE solve functionality using the matrices $A,C,M,N$ relevant to the RIEKF$^*$.  As a result, the gains $K$ will be based on  systems' noise parameters. The time varying components of the filter matrices are omitted when computing the steady state gain values, i.e., $\hat{I}_{\omega}=0$. %The experiment performed in Fig.\ref{fig:ahrs_gains2} is repeated for this modified NCF. 
Fig.\ref{fig:ahrs_gains2} (c) illustrates the gains of the filter found using this proposed approach which are equivalent to the steady state gains converged by the RIEKF$^*$ for the same trajectory.

The above process cannot be performed on the LIEKF$^*$ which in its noiseless form is equivalent to the generic NCF proposed by \cite{Mahony2008}. It is evident that despite the popularity of the NCF, its stochastic equivalent (i.e., LIEKF$^*$ shown in  this paper) does not allow one to identify a constant set of optimal steady state gains due to the existence of dominant coupling between the optimal gains and the system trajectory. On the other hand, the right invariant form of the NCF has a stochastic equivalent which converges to a constant set of dominant steady state gains, thereby allowing intuitive tuning based on the noise parameters of the system. Therefore, this paper proposes the steady state RIEKF$^*$ as the natural extension of the NCF, which by design inherits similar characteristics as the original NCF, but with the added intuitive tuning capability beneficial for practical applications. We term this formulation the RINCF.  The filtering equations of the RINCF are summarized in Table \ref{tbl:RINCF_eqns}, where the operator $Dare(\cdot)$ is used to denote the application of DARE to a set of matrices.
\begin{table}
\centering
\begin{tabular}{|c|}
\hline
    {$\begin{array}{l}
     ~~~~~~~~~~~~~~~~~~~~~~~~~~~~~~~~~~~~~~~~RINCF\\
        \mathrm{Initialization:}\\
            ~~~~\begin{array}{c c c}
            \hat{\mathbf{x}} =~\begin{bmatrix}1 &\mathbf{0} &\mathbf{0} \end{bmatrix}^\mathsf{T}
            &Q                =~ \mathrm{E}(\mathbf{w}\mathbf{w}^\mathsf{T})
            &R                =~\mathrm{E}(\boldsymbol{\nu}\boldsymbol{\nu}^\mathsf{T})\\
            \end{array}\\
        \mathrm{Filter Equations:}\\
        \begin{array}{c}
        \begin{array}{c }
            \begin{pmatrix}
                \dot{\hat{\mathbf{q}}}\\
                \dot{\hat{\omega}}_b
            \end{pmatrix}
                =
           \begin{pmatrix}
               \frac{1}{2}\hat{\mathbf{q}}*(\boldsymbol{\omega}_m-\hat{\boldsymbol{\omega}}_b )\\
               \mathbf{0}
            \end{pmatrix}
             +
            \begin{pmatrix}
               L_q\mathcal{E}*\hat{\mathbf{q}}\\
               \hat{\mathbf{R}}^{\mathsf{T}}_\mathbf{q}L_{\omega_b}\mathcal{E}\\
               %\hat{\mathbf{q}}^{-1}*L_{\omega_b}\mathcal{E}*\hat{\mathbf{q}}\\
            \end{pmatrix}\\
             \begin{pmatrix}
                \hat{\mathbf{y}}_a\\
                \hat{\mathbf{y}}_b\\
            \end{pmatrix}
                =
             \begin{pmatrix}
               -\hat{\mathbf{R}}_{\mathbf{q}}^\mathsf{T}\mathbf{g}_e\\
               \hat{\mathbf{R}}_{\mathbf{q}}^\mathsf{T}\mathbf{b}_e
             \end{pmatrix} 
             ~~~~~~~\mathcal{E} =~    \begin{pmatrix}
                             \hat{\mathbf{R}}_\mathbf{q}(\mathbf{\hat{y}_a}\times\mathbf{y}_{a})\\
                             \hat{\mathbf{R}}_\mathbf{q}(\mathbf{\hat{y}_b}\times\mathbf{y}_{b})\\
                             \end{pmatrix}\\
       %     & \mathrm{\emph{State Prediction}} & \mathrm{\emph{State Correction}} \\%& \mathrm{\emph{Output Prediction}}\\
        \end{array}\\
        \begin{array}{c}
            %\mathrm{\emph{Output Error}}~~
%            \mathcal{E} =~    \begin{pmatrix}
%                             \hat{\mathbf{R}}_\mathbf{q}(\mathbf{\hat{y}_a}\times\mathbf{y}_{a})\\
%                             \hat{\mathbf{R}}_\mathbf{q}(\mathbf{\hat{y}_b}\times\mathbf{y}_{b})\\
%                             \end{pmatrix}\\
            %&\mathrm{\emph{Gain Matrix}}~~K       =\begin{bmatrix}
             %               -L_q &-L_{\omega_b}
              %              \end{bmatrix}^\mathsf{T}
             % \mathrm{\emph{Gain Computation}}\\
             K=\begin{bmatrix}
             -L_q &-L_{\omega_b}
              \end{bmatrix}^\mathsf{T}
         \end{array}\\
            \end{array}\\
        \mathrm{Filtering~Matrices(used~to~compute~gain~K):}\\
            ~~~~~~~~~~~~~~~
             \begin{array}{l l}
            A=~\begin{bmatrix}
                \mathbf{0}   &-\frac{1}{2}\mathbf{I}\\
                \mathbf{0}   &\mathbf{0}
             \end{bmatrix}
            &C=~\begin{bmatrix}
                2[\mathbf{g}_e]_{\times}^2  &\mathbf{0}\\
                2[\mathbf{b}_e]_{\times}^2  &\mathbf{0}
             \end{bmatrix}\\
             M=~\begin{bmatrix}
                0.5\mathbf{I}  &\mathbf{0}\\
                \mathbf{0}  &-\mathbf{I}
             \end{bmatrix}
             &N=~\begin{bmatrix}
                \mathbf{I}+[\mathbf{g}_e]_{\times}  &\mathbf{0}\\
                \mathbf{0}               &\mathbf{I}-[\mathbf{b}_e]_{\times}
             \end{bmatrix}\\
            \end{array}\\
            ~~~~~~K=Dare(I+Adt,~C,~MQM^\mathsf{T}dt^2,~NRN^\mathsf{T})
            \end{array}
    $}
    \\
    \hline
\end{tabular}
\caption{The proposed right invariant nonlinear complementary filter}\label{tbl:RINCF_eqns}
\end{table}

%The resulting filter is low cost, and easily tunable using the system noise parameters off line. I.e., one can easily find K by using Matlab's discrete algebraic Riccatti equation (\textbf{DARE}) solving function with the 4 matrices given above as parameters. The resulting steady state gains are illustrated in  Fig.\ref{fig:ahrs_gains2}, where the RINCF was capable of achieving the optimal steady state dominant gains achieved by the RIEKF$^*$.

It is important to note that at high angular velocities some elements of the optimal gain matrix  deviates from its steady state values. By analyzing this behaviour it was identified that the gain matrix approximates to the following function with 2 extra parameters relevant to high angular velocity cases. 

%modified form
\begin{equation}\label{eqn:rincf2}
    \begin{array}{c c}
    K=
     \begin{pmatrix}
                  L_{q}\\
                  L_{\omega_b}
      \end{pmatrix} =
      \begin{pmatrix}
                 \mathbf{I}_{\mathbf{a}} &\mathbf{I}_{\mathbf{b}}\\
                 -\mathbf{I}_{\mathbf{c}}+p_1[\hat{-\mathbf{I}}_{\omega}]_{\times} &-\mathbf{I}_{\mathbf{d}}+p_2[\hat{-\mathbf{I}}_{\omega}]_{\times}
      \end{pmatrix}
      \end{array}
\end{equation}

An approximation for parameters $p_1$ and $p_2$ can be found by applying DARE solving function with $\hat{\mathbf{I}}_{\omega}=(\omega_{max},~0,~0)$. The parameters $p_1$ and $p_2$ can be calculated using the resulting gain matrix as $p_1=K_{6,2}/\omega_{max}$, $p_2=-K_{5,6}/\omega_{max}$. The scalar $K_{i,j}$ is used to denote the gain value located at column $i$, and row $j$ of gain matrix $K$.  The RINCF variant which has the gain function (\ref{eqn:rincf2}) is defined as RINCF2. The performance of RINCF2 is compared with RINCF for different experimental runs in the results section.

\section{Results}

This section evaluates the numerical and experimental performance of the RINCF for multitude of scenarios and compares the performance agaainst generic filter designs used for orientation estimation .

\subsection{Numerical validation of the RINCF}

Three trajectories were used for numerical evaluation of the proposed RINCF as given in equation set (\ref{eqn:trajs}). The trajectories were randomly selected to evaluate the performance of the proposed filter and its tuning procedure for low, medium, and high angular velocity cases.

\begin{equation}\label{eqn:trajs}
\begin{array}{l l}
\vspace{0.1cm}
Case~1:  &  
	\pmb{\omega}=\begin{pmatrix} \pi/3\sin(2\pi 0.7t+\pi/3)\\
            		         \pi/3\sin(2\pi 0.2t+\pi) \\
            			\pi/3\sin(2\pi 0.4t) 
             			\end{pmatrix} rad/s\\
\vspace{0.1cm}
Case~2:  &  
\pmb{\omega}=\begin{pmatrix} \pi\sin(2\pi 0.7t)\\
            			\pi\sin(2\pi .02t+\pi) \\
            			\pi\sin(2\pi .04t+\pi/3)
             			\end{pmatrix} rad/s\\
Case~3:  & 
	\pmb{\omega}=\begin{pmatrix} 5\pi/3\sin(2\pi .07t+\pi/3)\\
    				5\pi/3\sin(2\pi .02t+\pi)\\
    				5\pi/3\sin(2\pi .04t)\\
             			\end{pmatrix} rad/s\\	
\end{array}
\end{equation}

The system noise parameters were taken as $Q=0.1\mathbf{I}_6$, and $R=diag(0.3\mathbf{I}_6,0.5\mathbf{I}_6)$. The proposed RINCF was implemented with the gain matrix found through DARE runs in Matlab. The resulting tuning parameters using the proposed RINCF design method were $(a_1,~a_2,~b_2,~b_3,~c_1,~c_2,~d_2,~d_3)$ $=(0.3326$,$0.2517$,  $0.1511$, $0.2630$, $0.5666$, $0.4412$, $0.2648$, $0.4332)10^{-3}$.  These gain values were used for the RINCF in all of the numerical experiments.
%roll pitch yaw
\begin{figure}[h]
\centering
  \includegraphics[width=3.2in]{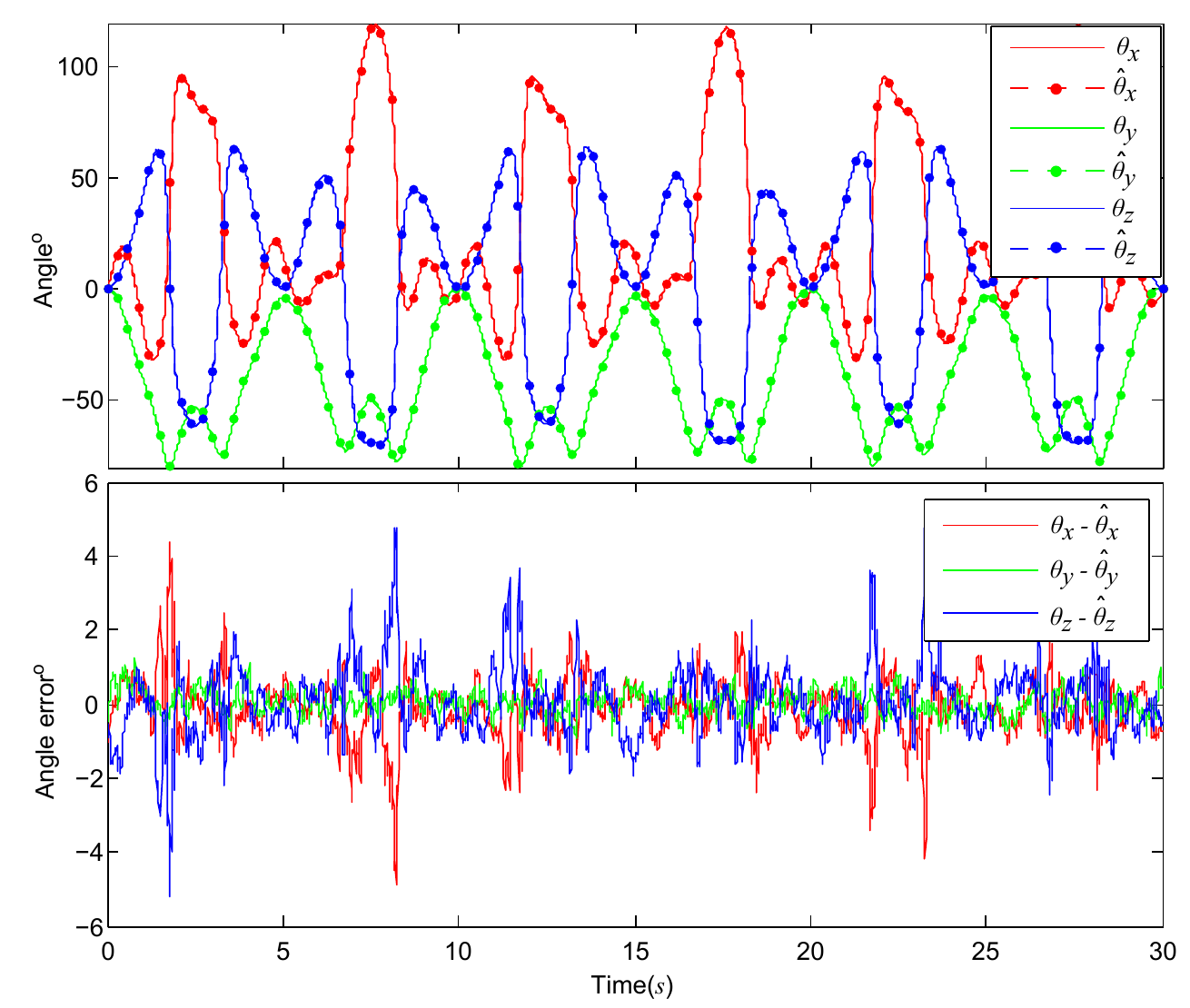}\\
  \caption{The simulated trajectory of the platform and the estimation accuracy provided by the RINCF for $Case ~1$}\label{fig:rincf_track}
\end{figure}

The results of the RINCF was compared against main attitude estimators reported in literature \cite{Markley1988,Farrell2008a,Martin2009,Mahony2008}. $Case~1$ was selected as the test case for this comparative analysis since the trajectory corresponds to typical operating velocities of aerial platforms. Fig.\ref{fig:rincf_track} illustrates the estimation of  roll($\theta_x$), pitch($\theta_y$), and yaw($\theta_z$) for $Case$ 1. The RINCF was capable of fast convergence and accurate estimation of attitude and heading using the proposed tuning approach.
%bar plot all
\begin{figure}[h]
\centering
  \includegraphics[width=3.2in]{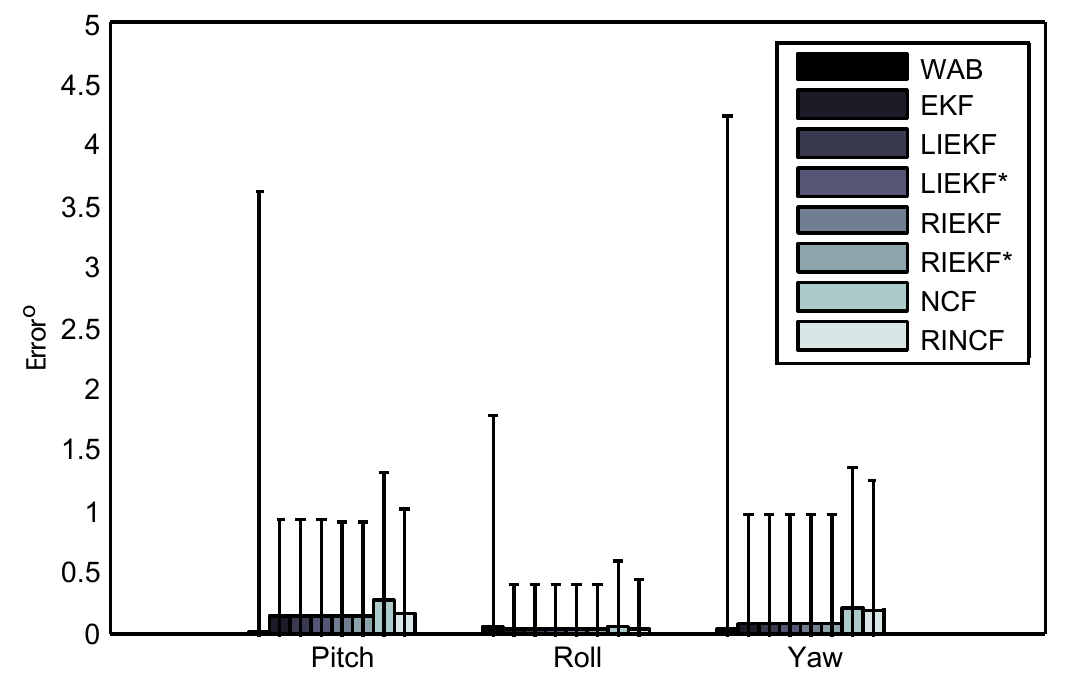}\\
  \caption{Performance of the RINCF for $Case ~1$ compared with main attitude estimators reported in this work}\label{fig:ahrs_result_all}
\end{figure}

\begin{figure*}[t!]
  \centering
  \includegraphics[width=5.5 in]{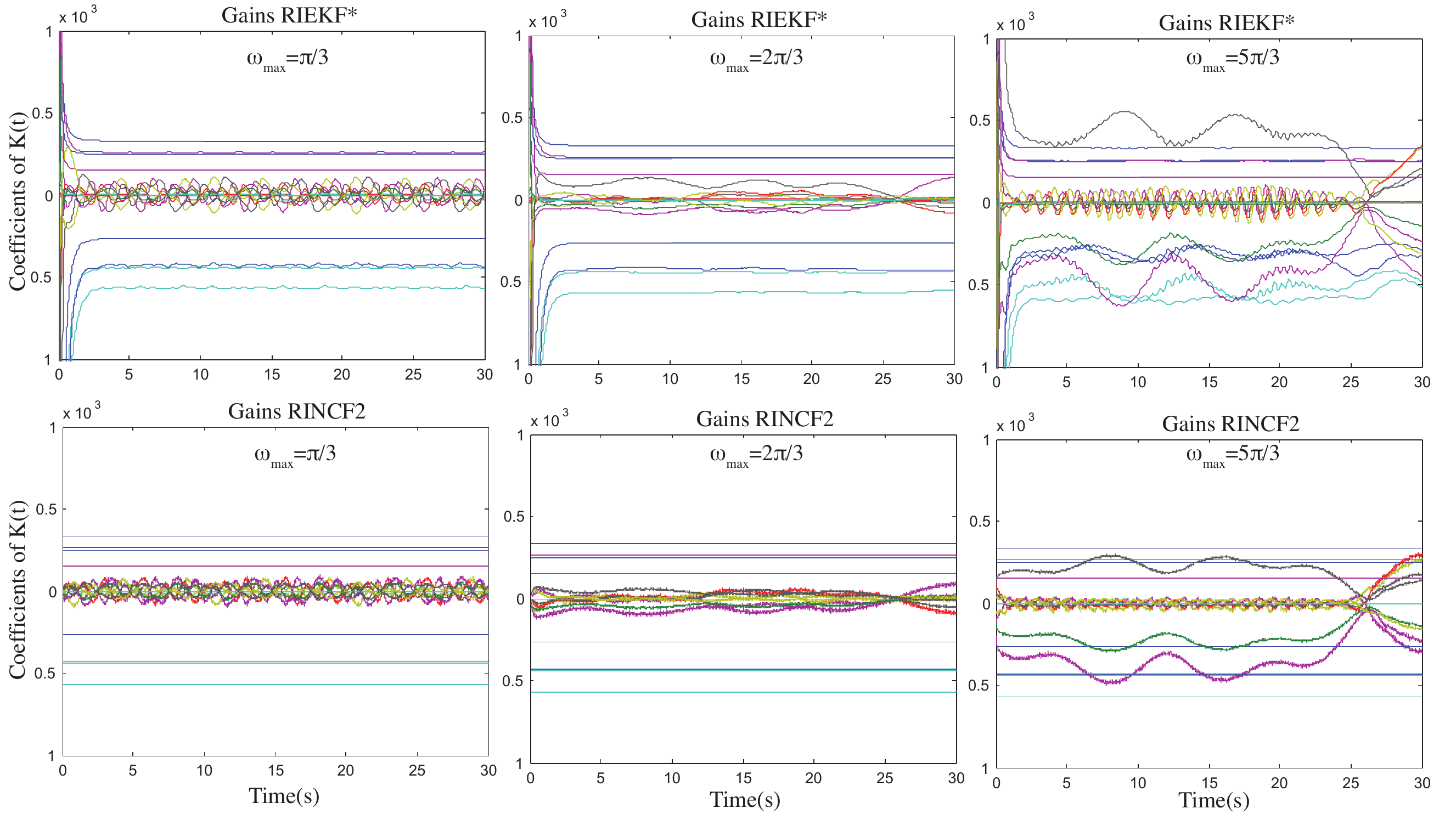}\\
  \caption{Gain matrix $K(t)$ of the RIEKF$^*$ (top) and the RINCF2 (bottom), for a smooth trajectory with maximum angular velocity $\omega_{max}=\pi/3$ (left), $\omega_{max}=2\pi/3$ (middle), and $\omega_{max}=5\pi/3$ (right)}\label{fig:ahrs_gains3}
\end{figure*}

%bar plot all cases
\begin{figure*}[t!]
  \centering
  \includegraphics[width=6.8in]{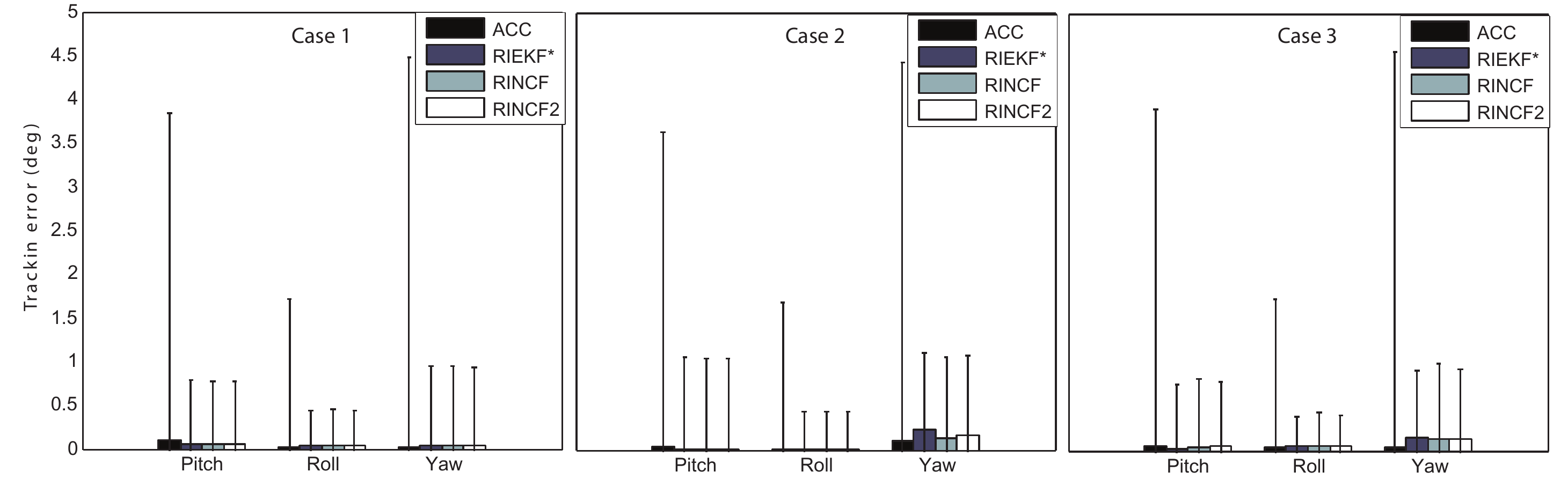}\\
  \caption{Comparative performance of the RINCF for all cases}\label{fig:rincf_all_cases}
\end{figure*}

Fig.\ref{fig:ahrs_result_all} illustrates the mean and standard deviation of estimation error produced by the filters after convergence (results after $2.5$ seconds). The filter denoted as $WAB$ is the solution achieved by solving the Wahba's problem using singular value decomposition \cite{Markley1988}. This denotes the attitude and heading solution achievable solely using an accelerometer and a magnetometer, thus corresponds to the worst case performance expected from the filters.  The RINCF provides comparable accurate results as all other estimators considered in this work as illustrated in Fig.\ref{fig:ahrs_result_all}. Out of the filters presented in Fig.\ref{fig:ahrs_result_all}, the NCF and the RINCF are the only deterministic implementations applicable with minimal computational overhead, while the RINCF dominates as the only deterministic filter which has system noise based gain tuning capability as proposed in this work. The gains of the NCF were selected by trial and error for this experiment.

\subsection{Higher angular velocity cases}

Next the performance of RINCF for higher angular velocity cases ($Case~2$ and $Case~3$) was evaluated. During higher angular velocity excitations the optimal gain matrix has elements which are significantly coupled with the trajectory. For this analysis the gains produced by RIEKF$^*$ are considered as optimal steady state gains for a given trajectory.  As illustrated in Fig.\ref{fig:ahrs_gains3}, RINCF2 have both dynamic and constant gain values which are comparable to optimal gains for all three cases. Whereas RINCF is only capable of having the constant optimal gain values. Fig.\ref{fig:rincf_all_cases} illustrates a comparison of estimation performance of RIEKF$^*$, RINCF and RINCF2 for the three cases.  It is evident from Fig.\ref{fig:rincf_all_cases}  that both RINCF and RINCF2 achieve comparable performance as the RIEKF$^*$ after convergence for cases with increasing angular rates. Although RINCF2 is able to produce better gains for higher angular velocity cases, it only resulted in marginal improvement in terms of estimation accuracy when compared with the constant gain matrix proposed in RINCF.  Therefore RINCF is recommended even for higher angular velocity excitations since 1) the filter produces accurate estimation comparable to stochastic forms, 2) the filter is easier to tune with less parameters than RINCF2 and 3) the filter guarantees local asymptotic stability as shown in Section \ref{sec:rincf}.

%it is evident from Fig.\ref{fig:rincf_all_cases} that even without the modifying terms the proposed RINCF can be used for accurate orientation estimation with guaranteed local asymptotic stability. %The results validate the proposed RINCF design for attitude and heading estimation and the proposed tuning method which allows to achieve performance comparable to a steady state stochastic filter over a wide range of system configurations and trajectories. 

\subsection{Experimental validation of the RINCF}

\begin{figure}[h]
  \centering
  \includegraphics[width=3.2in]{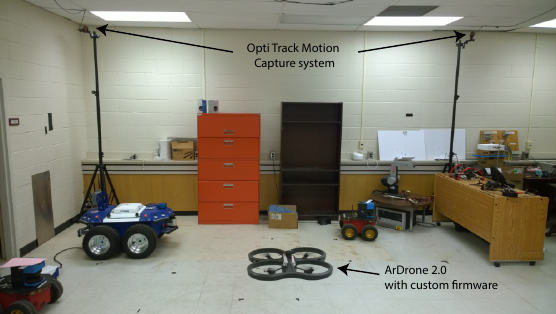}\\
  \caption{The experimental validation setup used in the study}\label{fig4:ardrone_mag_calib}
\end{figure}

The experimental validation of the RINCF was performed using an ARDrone 2.0 MAV platform. The MAV was programmed with custom firmware which records the raw measurements from the onboard accelerometer, gyroscope, and magnetic compass. Scale factors and bias values related to these sensors were identified. Additionally the hard iron distortion effect of the digital compass was identified and calibrated prior to experiments. These calibrated measurements were used to implement each type of filter. The performance was evaluated using the attitude and heading estimates provided by an OptiTrack motion capture system when the platform moved along a random trajectory. The RINCF was implemented on the MAV's onboard processor to validate its capability to operate with minimal computational overhead. The filtering equations required only $156 \mu s$ to complete an execution cycle which represents less than  $4\%$ of the $200Hz$ program execution rate of the MAV.

%\begin{figure}[h]
%  \centering
%  % Requires \usepackage{graphicx}
%  \includegraphics[width=3.2in]{ardrone_mag_calib2}\\
%  \caption{Calibration of the magnetic compass of the ARDrone 2.0}\label{fig4:ardrone_mag_calib}
%\end{figure}

It is important to note that the magnetic heading of the MAV is highly biased for indoor operating locations. These errors have been minimized by calibrating the compass for the test location. This is the case even when the drone's  yaw is to be controlled in a closed loop using the factory firmware. In this case the factory software performs a $360^0$ rotating data capture from the magnetic compass and identifies the hard iron distortion parameters in order to operate in the particular environment. 

\begin{figure}[h]
  \centering
  \includegraphics[width=3.2in]{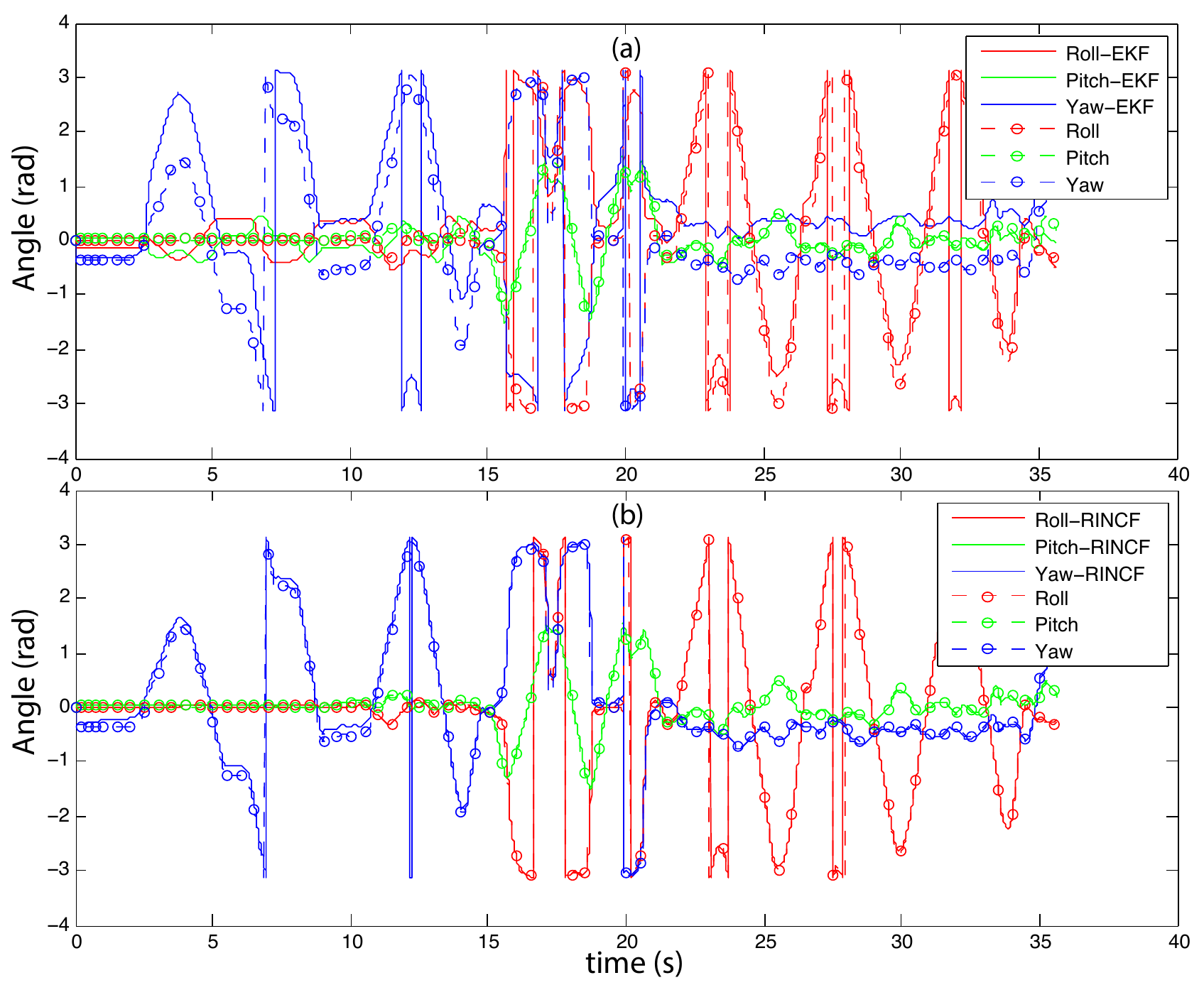}\\
  \caption{(a)Attitude heading estimates of the EKF, (b) Attitude heading estimates of the RINCF for experimental data}\label{fig4:ardrone_ekf_results}
\end{figure}

Fig.\ref{fig4:ardrone_ekf_results}(a) illustrates the performance of an EKF for measurements captured by the MAV. The generic EKF was implemented with an additive noise assumption, and the results exhibit poor estimation capability. This is mainly due to the biases of the magnetic sensor which ultimately affects roll and pitch estimates due to dominant coupling effects. In contrast the RINCF estimates illustrated in Fig.\ref{fig4:ardrone_ekf_results}(b) show better performance due to minimal coupling of the magnetic compass with the roll and pitch estimates. The right invariant formulation of the filter allows to manipulate the gain matrix in order to force the magnetometer to primarily affect the yaw estimate while minimizing its effect on roll and pitch estimates. Equation (\ref{RINCF_gain}) reports the gain values estimated using the noise parameters related to the sensors. In order to force the desired selective updating by the magnetometer, it is necessary to set the elements of the gain matrix $K_{RINCF}$, which are highlighted in bold  to zeros. All other values are calculated using the discrete algebraic Ricatti solve function of Matlab. It was not necessary to introduce the additional $p_1$ and $p_2$ terms in calculating the gain matrix. Only the RIEKF and RINCF allowed the convenience of incorporating the selective update  capability using the magnetometer. The EKF and the NCF only allowed to manipulate the confidence between the accelerometer and the compass; and did not allow selective updating of states similar to right invariant formulations.

\begin{equation}\label{RINCF_gain}
    \begin{array}{c c}
    K_{RINCF} &= \begin{pmatrix}  	 -0.0037   &0       	&0    &\mathbf{0}    &0  &0\\
                       						 0    &-0.0037     &0    &0   &\mathbf{0}   &0\\
                        				 		 0    &0   &\mathbf{0}    &0    &0    &-0.001\\
                       					0.0028   &0             &0    &\mathbf{0}    &0   &0\\
                       				 	0              &0.0028   &0    &0   &\mathbf{0}   &0\\
                       					0   &0   &\mathbf{0}             &0   &0   &0.0002 \end{pmatrix} 
					
%   -0.0037   -0.0000         0   -0.0000    0.0000         0
%   -0.0000   -0.0037         0    0.0000   -0.0000         0
%         0         0         0         0         0   -0.0010
%    0.0028    0.0000         0    0.0000   -0.0000         0
%    0.0000    0.0028         0   -0.0000    0.0000         0
%         0         0         0         0         0    0.0002
					
    %Case~1 & Case~2 & Case~3\\
    %\omega_{x}=\pi/3\sin(2\pi 0.7t+\pi/3)   &\omega_{x}=\pi\sin(2\pi 0.7t)          &\omega_{x}=5\pi/3\sin(2\pi .07t+\pi/3)\\
    %\omega_{y}=\pi/3\sin(2\pi 0.2t+\pi)     &\omega_{y}=\pi\sin(2\pi .02t+\pi)    &\omega_{y}=5\pi/3\sin(2\pi .02t+\pi)\\
    %\omega_{z}=\pi/3\sin(2\pi 0.4t)         &\omega_{z}=\pi\sin(2\pi .04t+\pi/3)        &\omega_{z}=5\pi/3\sin(2\pi .04t)
    \end{array}
\end{equation}
% K_ss =1.0e-03 *[   0.0229   -0.0000         0    0.0000    0.0000   -0.0000;
%                       -0.0000    0.0228         0    0.0000    0.0000    0.0000;
%                        0.0000   -0.0000         0   -0.8837    0.5025    0.4929;
%                       -0.0202    0.0000         0   -0.0000   -0.0000    0.0000;
%                        0.0000   -0.0201         0   -0.0000   -0.0000   -0.0000;
%                       -0.0000    0.0000         0    0.1741   -0.0990   -0.0971];

\begin{figure}[h]
  \centering
  \includegraphics[width=3.2in]{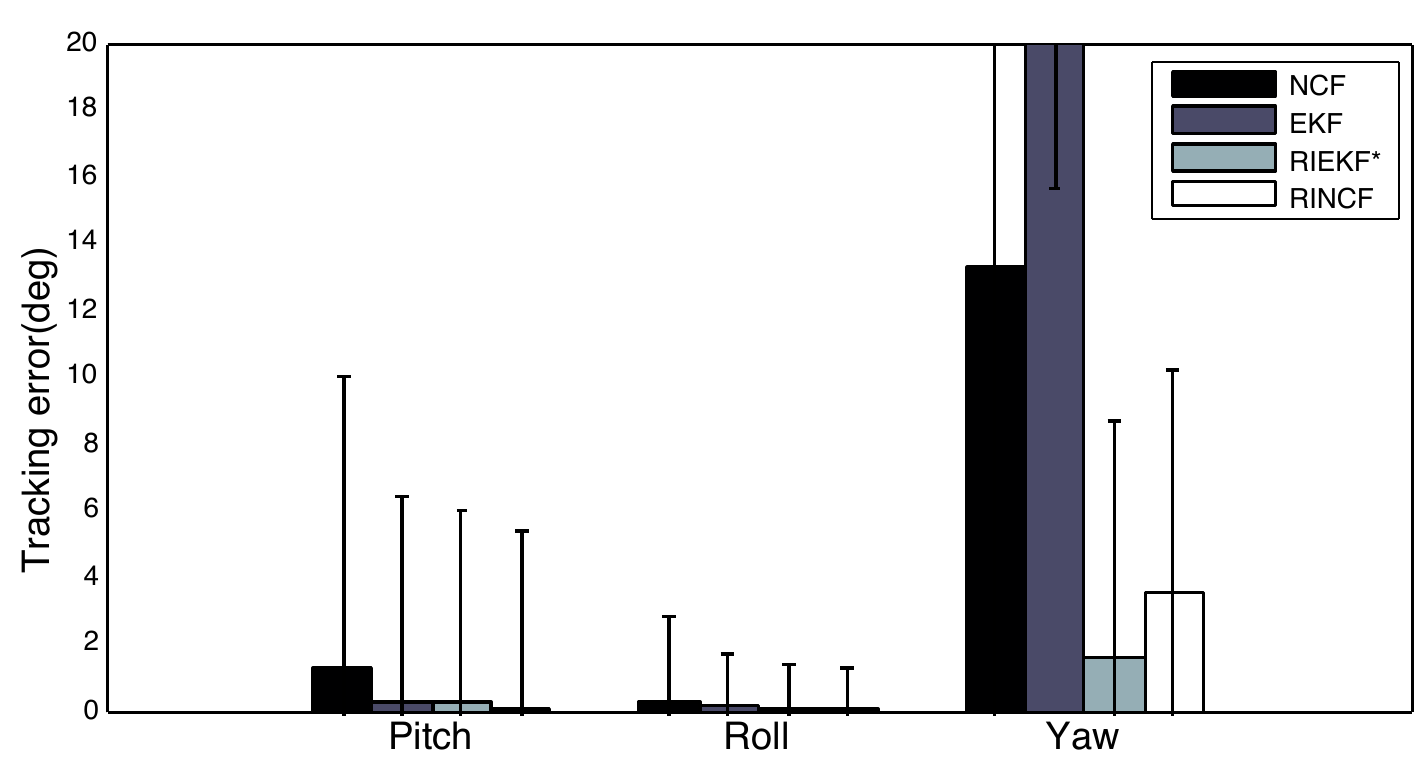}\\
  \caption{Attitude heading estimates of the filters for experimental data. RINCF performance was validated by implementing the filter on the MAV platform.}\label{fig4:ardrone_results_all}
\end{figure}

Experimental results related to an EKF, NCF, RIEKF, and an RINCF are presented in Fig.\ref{fig4:ardrone_results_all}.  The results illustrate accurate estimation of the RINCF with comparable performance as its stochastic form when implemented on MAV platforms. Therefore, the RINCF constitutes an excellent novel approach for the attitude filtering problem which has the added capability of tuning the filter using an intuitive approach, and the capability of deploying on devices with low computational demand.

\section{Conclusion}

This paper proposed a novel formulation of the nonlinear complementary filter termed the RINCF, which has the ability to be implemented as a low cost deterministic AHRS with a set of fixed gain values. The novelty of RINCF is that it possesses an intuitive method of tuning its gains which bases on the system's noise parameters. Hence considered a significant improvement over the generic NCF for attitude estimation. The proposed RINCF achieves performance comparable to its equivalent stochastic implementation at steady state. The paper presents a detailed formulation of the filter, its stability analysis, and validation of the design using both numerical and experimental results. The filter and the proposed tuning method was successfully deployed on an MAV platform as part of the experimental validation process where accurate orientation estimation was achieved with minimal computational overhead. 

Future work related to this design attempts to further analyze the robustness of the filter for practical applications with higher angular velocities and magnetometers with bias fluctuations. The methodology proposed in this work to arrive at a steady state filter using an IEKF formulation, is being further exploited in other filtering domains such as relative localization and inertial navigation systems to realize low cost implementations  with intuitive gain tuning capability.

%The work attempts to analyze stability implications related to modifications such as extra parameters in the gain function and forcing magnetometers to only correct yaw estimates. Furthermore, this filter is being modified to include new developments in AHRS design for MAV platforms reported in \cite{Abeywardena2013a} \cite{Leishman2014}. The capability allows to incorporate the relationship between the platform velocities and the acceleration measurements recorded along the body plane of the platform, in the filter design process.  The methodology proposed in this work to arrive at a steady state filter using an IEKF formulation, is being further exploited in other filtering domains such as relative localization and inertial navigation systems to realize low cost implementations  with intuitive gain tuning capability.

\bibliographystyle{asmems4}
\bibliography{sensors_mdpi}

\end{document}